\documentclass[letterpaper]{article}

\usepackage[preprint]{aaai2027}
\usepackage[hyphens]{url}
\usepackage{graphicx}
\usepackage{natbib}
\usepackage{caption}
\usepackage{amsmath,amssymb}
\usepackage{booktabs}
\usepackage{multirow}
\usepackage{xcolor}
\usepackage{array}
\usepackage{float}
\newcommand{\moc}{\mbox{\normalfont\scshape MoC}}

\newcommand{\R}{\mathbb{R}}
\newcolumntype{L}[1]{>{\raggedright\arraybackslash}p{#1}}
\newcommand{\lewm}{\mbox{\normalfont\scshape LeWM}}
\newcommand{\cem}{\mbox{\normalfont\scshape CEM}}
\newcommand{\asar}{\mbox{\normalfont\scshape ASAR}}
\newcommand{\trm}{\mbox{\normalfont\scshape TRM}}
\newcommand{\A}{\mathbf{A}}
\newcommand{\z}{\mathbf{z}}
\newcommand{\x}{\mathbf{x}}
\newcommand{\zg}{\mathbf{z}_{g}}
\newcommand{\aRec}{\mathbf{A}_{\mathrm{rec}}}

\title{Action from Adjacent Set in Physical Space\\
Outperforms the Best Prediction in World Models}

\author{Liangyu Li, Qingwen Liu, Mingqing Liu, Wen Fang}
\affiliations{Tongji University, Shanghai, China\\
Liangyu Li: \texttt{2550703@tongji.edu.cn}\\
Qingwen Liu: \texttt{qliu@tongji.edu.cn}\\
Mingqing Liu: \texttt{clare@tongji.edu.cn}\\
Wen Fang: \texttt{wen.fang@tongji.edu.cn}}

\begin{document}
\maketitle

\begin{abstract}
Controllers based on sampling and latent world models assign a predicted terminal cost to each candidate action sequence, choose the minimum, execute its first action block, and replan. This rule can fail even when the terminal cost perfectly and accurately reflects the true task objective in the physical world. Residual prediction error can give an infeasible sequence an anomalously low cost, and a larger proposal pool gives such errors more chances to outrank feasible alternatives. We call this conditional failure
\emph{proposal overgeneration}. In Cube candidate execution audits, increasing the total proposal budget from 72 to 288 reduces the feasibility of selection by minimum latent cost from .375 to .062 for position targets and from .344 to .031 for targets defined by position and yaw, although every larger pool contains a feasible sequence. We introduce Adjacent Set Action Reconstruction (\asar{}). Among proposals with low cost, \asar{} identifies an adjacent set using standardized early action prefixes and reconstructs a full action sequence through locally weighted aggregation with a light anchor from the sequence with minimum cost. On a Carry and Release evaluation set of 75 queries, Kernel \asar{} improves event completion success over matching selection by 28.0, 24.0, and 18.7 percentage points under latent cost and by 18.7, 20.0, and 17.3 points under a trajectory reachability cost at 72, 144, and 288 proposals. Analysis of finite proposal pools characterizes selection risk from the lower tail, separation by a related radius support statistic, and sequence containment under an explicit local feasibility condition.
\end{abstract}

\section{Introduction}

Control based on sampling and latent world models encodes the current and goal observations, predicts the terminal latent state for each candidate action sequence, and ranks the candidates with a terminal cost. A receding-horizon controller then executes the first action block of the selected sequence and replans. This pattern appears in latent model predictive control and recent world models for visual planning~\citep{hafner2019planet,hansen2022tdmpc,
hansen2024tdmpc2,hafner2024dreamerv3,micheli2023iris,pu2025unizero,
zhou2024dinowm,assran2025vjepa2,maes2026leworldmodel}. Increasing the proposal budget improves candidate coverage, and a better terminal cost improves the evaluation of predicted endpoints. Neither operation explicitly uses distances between the action prefixes in the current proposal pool.

\begin{figure*}[t]
\centering
\includegraphics[width=0.98\textwidth]{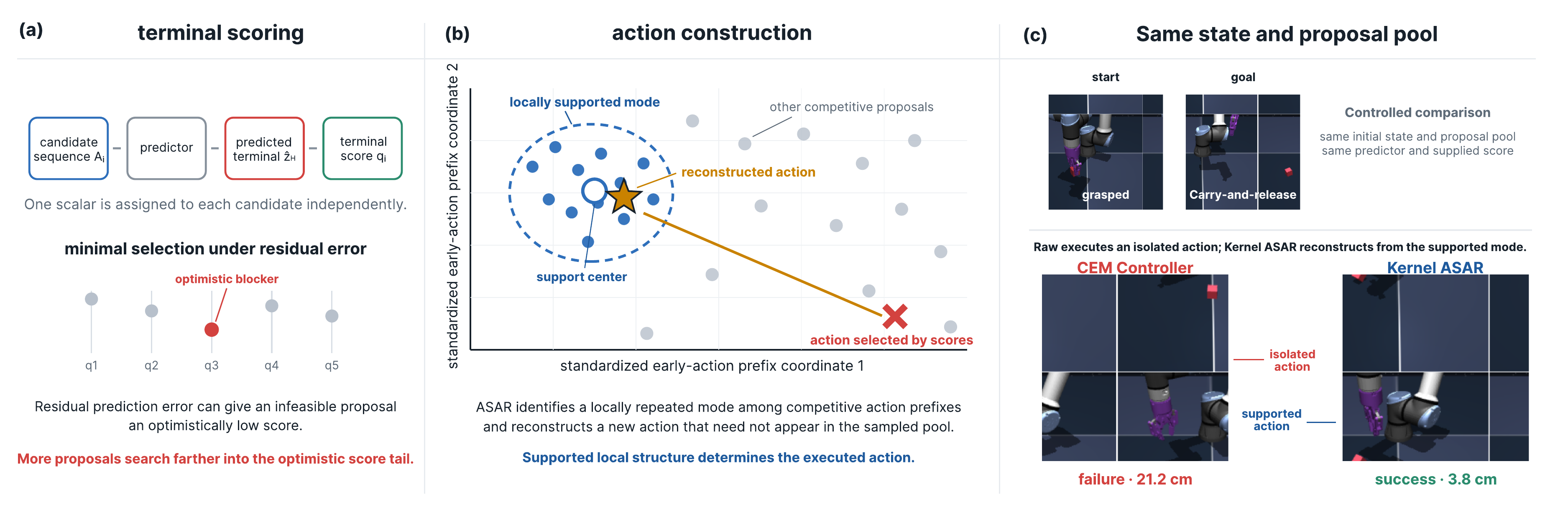}
\caption{From terminal scoring to adjacent set action reconstruction. (a) A
fixed predictor assigns a terminal score to each candidate independently, so
residual prediction error can place an infeasible proposal in the optimistic
score tail. (b) \asar{} identifies a locally repeated mode among competitive
early action prefixes and reconstructs a new full action sequence from this
adjacent set. (c) In Carry and Release, the two controllers use the same state,
proposal pool, predictor, and terminal score. Minimum score selection executes
an isolated action and leaves the cube 21.2 cm from the target, while Kernel
\asar{} reconstructs from the adjacent set and completes the task within
3.8 cm.}
\label{fig:carry_release}
\end{figure*}

Figure~\ref{fig:carry_release} shows why these relations within the pool can matter.
The controller begins with a grasped cube and must transport it to a lower target before release.  The sequence with minimum latent cost has an isolated prefix in the full standardized prefix space.  After its first action block is chosen at each replan, the cube finishes 21.2 cm from the target.  Twelve other sequences with low cost have nearby prefixes.  Kernel \asar{} reconstructs a weighted full action sequence from that adjacent set and the sequence with minimum cost.  The resulting closed-loop execution releases the cube within 3.8 cm of the target from the same initial state and matched proposal draws.

The failure begins when the controller minimizes costs computed from predicted terminal states rather than terminal states reached after execution.  A useful predictor may still assign an infeasible sequence a cost below every feasible alternative. When more infeasible sequences are sampled from a comparable
conditional distribution, the probability of observing at least one such error in the lower tail increases. Feasible sequence coverage can therefore improve while the fraction of pools with a feasible sequence among the lowest costs decreases.  We refer to this budget-induced degradation in candidate ranking under residual model error as proposal overgeneration.

We measure this pattern in simulated Cube manipulation with a fixed \lewm{} checkpoint and \cem{} controller. In an audit of 192 proposal pools where every candidate is executed independently from the same state, selection by minimum latent cost chooses an infeasible sequence in 132 pools. Among these 132 failures, 58 pools, or 43.9\%, already contain a feasible alternative among the 80 lowest latent costs. A separate evaluation set contains 32 pools for position targets and 32 for targets defined by position and yaw. Increasing the total budget from 72 to 288 lowers the fraction of pools with a feasible sequence among the 20 lowest costs from .969 to .656 and .625. A trajectory reachability cost raises the two rates at 288 proposals to .750 and .719, but the mean number of infeasible sequences with lower cost still increases substantially.

These results motivate an adjacent-set action reconstruction rule. When a region of low cost in the space of full action sequences has a robust feasible neighborhood, nearby early prefixes provide a proxy available at planning time for that local regularity.  \asar{} first restricts attention to sequences with low cost. It measures each sequence by the mean distance to its nearest standardized prefixes, identifies an adjacent set around the least isolated prefix, and reconstructs a weighted full action sequence from that set. A light mixture with the sequence of minimum cost retains information from the task objective. The world model, proposal generator, and terminal cost are unchanged. Selecting the least isolated existing sequence from the same set of sequences with low cost is the matched existing-candidate baseline without reconstruction.

Terminal objective quality and residual prediction error occur at different stages. Li et al. introduced Trajectory Reachability Metrics (\trm{}) to examine whether terminal latent distance reflects physical task progress when evaluating a predicted endpoint and to replace that distance with a trajectory-based metric~\citep{li2026trm}.
\asar{} begins after a terminal cost has been chosen. It assumes the cost is a meaningful task objective, then addresses cases where residual endpoint prediction error gives one sequence a misleadingly low cost. We therefore use latent distance and \trm{} as two compatible cost inputs, not as competing controllers, and compare minimum cost selection with adjacent-set action reconstruction under each cost.

Our contributions are:
\begin{enumerate}
    \item We identify proposal overgeneration under residual model error.  The candidate execution audits separate feasible sequence presence from top ranking and show that a larger pool can contain feasible sequences while minimum cost selection becomes less reliable.
    \item We propose \asar{}, which identifies an adjacent set among action prefixes with low cost and reconstructs a full action sequence that need not occur in the sampled pool. 
    \item We provide an order statistic identity for the lower tail, a concentration result under an explicit gap in radius support, and a conditional containment bound when the selected full sequences lie inside a robust feasible neighborhood.
    \item Across 72, 144, and 288 proposals, \asar{} improves event completion success under both latent distance and a reachability cost aligned with the task. Separate tests show a reconstruction gain of 13.7 percentage points over the least isolated existing sequence and replicate the method with independently trained world model weights.
\end{enumerate}

\section{Related Work}

\paragraph{Latent world model control.}
PlaNet established online planning through learned latent dynamics, while Dreamer and DreamerV3 optimize behavior in imagined trajectories
~\citep{hafner2019planet,hafner2020dreamer,hafner2024dreamerv3}. TD-MPC and TD-MPC2 combine compact prediction with receding-horizon control
~\citep{hansen2022tdmpc,hansen2024tdmpc2}. IRIS and UniZero study scalable planning in learned latent models~\citep{micheli2023iris,pu2025unizero}. Recent visual models plan over pretrained or jointly learned embeddings
~\citep{zhou2024dinowm,assran2025vjepa2,maes2026leworldmodel,
maes2026stableworldmodel}. Other work improves the planning interface or the model horizon~\citep{parthasarathy2025closing,du2026vlwm}. We leave the encoder, predictor, and action sampler fixed and use only relations among sequences in the final proposal pool.

\paragraph{MPC and \cem{}.}
\cem{} iteratively samples candidates, retains an elite set, and refits its proposal distribution~\citep{rubinstein1999crossentropy}. It is widely used with learned dynamics because it does not require differentiating through the environment~\citep{chua2018pets,hafner2019planet}. iCEM improves sample efficiency through colored noise sampling, elite reuse, and distribution constraints~\citep{pinneri2021icem}, while MPPI forms action updates weighted by cost~\citep{williams2017mppi}. DecentCEM maintains several independent search distributions to preserve multimodal structure~\citep{zhang2022decentcem}. Learned action priors, diffusion policies, and generators conditioned on subgoals also improve the distribution of candidate behavior~\citep{chi2024diffusionpolicy,wang2026prism,cheng2026sage}. These methods change candidate generation or the iterative distribution update.  \asar{} instead processes the final proposal pool, identifies an adjacent set in prefix space, and reconstructs a full action sequence from it.

\paragraph{Model error, objective quality, and reachability.}
Learned planners can select actions that exploit model error, and predictive accuracy under one loss need not imply accuracy for the planning objective
~\citep{janner2019mbpo,lambert2020objective}. Choosing the minimum among noisy estimates is also related to the optimizer's curse
~\citep{smith2006optimizerscurse}. Reachability methods replace geometric proximity with evidence about controllable progress from replay graphs, latent landmarks, temporal distances, or directed quasimetrics
~\citep{eysenbach2019sorb,zhang2021worldgraph,qian2023replan,bae2024tldr,
myers2024learningtemporal,myers2025quasimetric}. RC-aux was introduced to change world model training through multihorizon reachability supervision~\citep{li2026predictive}, while \trm{} was introduced to change the terminal objective for a predicted endpoint
~\citep{li2026trm}.  \asar{} addresses the subsequent residual failure. Given a meaningful terminal cost, it uses prefix distances within the pool to identify an adjacent set and reconstruct the sequence passed to receding-horizon execution.

\section{Definitions and Control Pattern}

Let $\x_t\in\mathcal{X}$ be the physical state at one replanning step and $o_t=g(\x_t)$ its observation.  An encoder $e$ maps the observation to $\z_t=e(o_t)$ and the goal observation $o_g$ to $\zg=e(o_g)$.  A candidate \emph{action sequence} is $\A_i=(\mathbf{a}_{i,0},\ldots,\mathbf{a}_{i,H-1})\in\mathcal{A}^{H}$, where one model action is an action block of repeated environment actions.
The learned predictor produces
\begin{equation}
    \hat{\z}_{i,H}=\hat f_H(\z_t,\A_i).
    \label{eq:rawcost}
\end{equation}
The controller assigns a terminal cost:
\begin{equation}
    q_i=Q(\hat{\z}_{i,H},\zg).
    \label{eq:supplied_score}
\end{equation}
Lower values are preferred.  In the latent cost condition,
$Q(\hat{\z},\zg)=\|\hat{\z}-\zg\|_2^2$.  In the reachability cost condition,
$Q$ is the trajectory-based cost from \trm{}~\citep{li2026trm}.  The two costs
are inputs to the same reconstruction rule.  In either case, $q_i$ is computed
from candidate $i$ without using the executed outcome of another candidate.

For a candidate execution audit, let $F_H$ denote the physical transition map
for the full horizon.  It reaches
$\x^{\mathrm{exec}}_{i,H}=F_H(\x_t,\A_i)$ and therefore
\begin{equation}
    \z^{\mathrm{exec}}_{i,H}=e\!\left(g(\x^{\mathrm{exec}}_{i,H})\right).
\end{equation}
The predicted terminal latent $\hat{\z}_{i,H}$ and the executed terminal
latent $\z^{\mathrm{exec}}_{i,H}$ are different objects.  The planner uses the
former.  The audit uses the latter only to label the candidate after an
independent execution.  In closed-loop control, only the first action block of
the selected sequence is executed before a new pool is generated.

Let $\ell_H(\x,o_g)$ include the terminal task conditions used by the audit,
including target error and, when required, the release condition.  For
tolerance $\epsilon$, the feasible set over the full horizon is
\begin{equation}
    \mathcal{E}_{\epsilon}(\x_t,o_g)
    =\{\A\in\mathcal{A}^{H}:\ell_H(F_H(\x_t,\A),o_g)\leq\epsilon\}.
\end{equation}
For a proposal pool $\mathcal{C}_B=\{\A_i\}_{i=1}^{B}$, write $Y_i=1$ when
$\A_i\in\mathcal{E}_{\epsilon}$.  A \emph{ranking blocker} under $Q$ is an
infeasible proposal $b$ with
\begin{equation}
    q_b < \min_{i:Y_i=1} q_i.
    \label{eq:blocker}
\end{equation}
The blocker count is the number of infeasible proposals with cost below that of the feasible proposal with the lowest cost. It is a diagnostic for selection by minimum cost, not a claim that every reconstruction rule is blocked by each such proposal.

In receding-horizon \cem{}, three fixed action noise scales produce a shared proposal pool. The world model predicts every candidate, and the elite set with the lowest costs updates the sampling distribution. After the last iteration, the controller executes the first action block of the selected sequence, observes the next image, and replans. We use \emph{total proposal budget} for the number of action sequences in one pool. A \emph{query} is one closed-loop evaluation from a start state to a goal state, which contains several replanning pools.

\section{Proposal Scaling under Model Error}

At total proposal budget $B$, let $P_B$ be the event that at least one
feasible sequence is present in the pool.  Let $T_{B,k}$ be the event that at
least one feasible sequence has rank at most $k$ under $q$.  The top-$k$
feasibility event satisfies
\begin{equation}
    \Pr(T_{B,k})=\Pr(P_B)\Pr(T_{B,k}\mid P_B).
    \label{eq:presence_survival}
\end{equation}
The first factor measures proposal coverage.  The second measures whether a
feasible sequence remains near the top after scoring.  The identity is exact;
the empirical question is how the two factors change with $B$, $Q$, and $k$.

\begin{figure*}[t]
\centering
\includegraphics[width=0.72\textwidth]{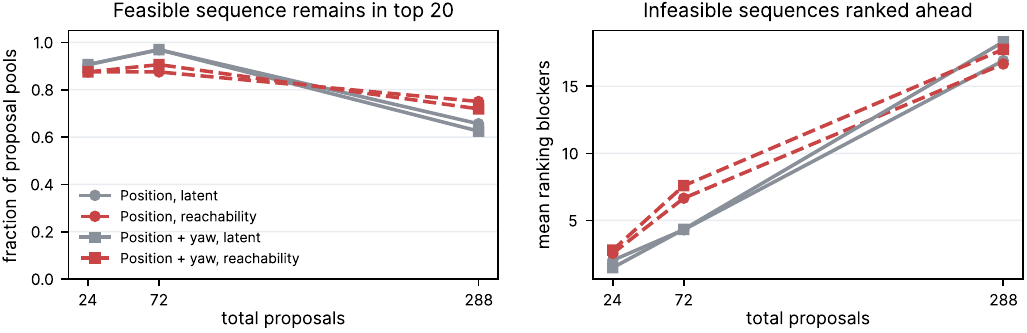}
\caption{Proposal scaling in the candidate execution audit. The fraction of pools with a feasible sequence among the 20 candidates with the lowest costs declines from 72 to 288 total proposals, although feasible presence is already one at both budgets. Mean counts of ranking blockers increase in the same comparison. The reachability cost changes the level of the curves but does not remove the effect of the larger pool.}
\label{fig:overgeneration}
\end{figure*}

We test this decomposition in a separate candidate execution evaluation set with
32 pools for each target family.  Every candidate in each pool is executed
independently from the same initial state to obtain $Y_i$.  The two cost rules
are then applied to the identical candidates.  Table~\ref{tab:overgeneration}
and Figure~\ref{fig:overgeneration} show that the reachability cost improves
top-20 coverage at the larger budget, while infeasible sequences with lower cost
remain.

\begin{table}[t]
\centering
\scriptsize
\setlength{\tabcolsep}{3pt}
\begin{tabular}{llrrrr}
\toprule
& & \multicolumn{2}{c}{Top-20 exposure} & \multicolumn{2}{c}{Mean blockers}\\
\cmidrule(lr){3-4}\cmidrule(lr){5-6}
Target family & Cost & $B=72$ & $B=288$ & $B=72$ & $B=288$\\
\midrule
Position & Latent & .969 & .656 & 4.31 & 16.91\\
Position & Reachability & .875 & \textbf{.750} & 6.66 & \textbf{16.66}\\
Position + yaw & Latent & .969 & .625 & 4.34 & 18.31\\
Position + yaw & Reachability & .906 & \textbf{.719} & 7.59 & \textbf{17.75}\\
\bottomrule
\end{tabular}
\caption{Candidate execution audit of proposal overgeneration. Each target family has 32 pools and every candidate is executed independently. At both total budgets, every pool contains at least one feasible sequence. Top-20 exposure is the fraction of pools with a feasible sequence among the 20 lowest costs. Blockers is the mean number of infeasible sequences with cost below that of the feasible sequence with the lowest cost.}
\label{tab:overgeneration}
\end{table}

The same pattern appears in closed-loop selection by minimum cost. From 72 to 288 total proposals, the feasibility of the selected sequence falls from .375 to .062 for position targets and from .344 to .031 for targets defined by position and yaw under latent cost. Under the reachability cost, the corresponding rates are .281 to .188 and .281 to .156. These three budgets are evidence for a non-monotone effect in this proposal distribution, rather than a claim that every increase in budget is harmful.

The mechanism in a finite pool can be stated for a conditional score distribution.
Condition on $n_-$ infeasible proposals and let
$F_{-}(t)=\Pr(q_b\leq t\mid Y_b=0)$.
Under conditional independence,
\begin{equation}
    \Pr\!\left(\min_{b:Y_b=0}q_b\leq t\right)
    =1-\left(1-F_{-}(t)\right)^{n_-}.
    \label{eq:lower_tail}
\end{equation}
For any $0<F_-(t)<1$, the probability increases strictly with $n_-$.  The
  identity says that the chance of sampling a low-cost infeasible sequence grows;
it does not say that the score error of an individual sequence grows.  A better
  cost can reduce $F_{-}(t)$, while residual probability in that tail remains
relevant at larger $n_-$.  The supplement gives an expectation expression for
blocker counts that does not require conditional independence.

The candidate execution audit connects this ranking pattern to physical outcomes. In 58 of 132 failures under latent cost, a feasible sequence is already among the 80 candidates with the lowest costs. In a separate diagnostic with 48 pools, replacing predicted terminal latents with latents encoded after independent execution raises minimum cost selection from .292 to .500, while ranking by the true terminal physical loss reaches .542. These are diagnostic upper bounds for selection over one horizon, not information available to the controller. Full accounting and the control with shuffled reachability costs appear in the supplement.

\section{Adjacent Set Action Reconstruction}

\subsection{Adjacent Set Identification}

The terminal cost evaluates candidates separately.  \asar{} adds one relation that is available only after the current pool has been generated: whether a prefix with low cost has nearby prefixes from other candidates. The intended setting is a locally regular action region, where small changes in a prefix are likely to preserve the same behavior over a short horizon.

Let $\mathcal{R}_M(Q)$ be the $M$ candidates with lowest terminal cost.  For
each candidate, vectorize the first $h$ model action blocks into an action
prefix $\mathbf{p}_i$ and robustly standardize each coordinate inside
$\mathcal{R}_M$.  Define the local prefix isolation score as the mean distance
to the $K$ nearest other prefixes,
\begin{equation}
    u_i=\frac{1}{K}\sum_{j\in\mathcal{N}_K(i)}
    \|\mathbf{p}_i-\mathbf{p}_j\|_2.
    \label{eq:support}
\end{equation}
The neighborhood excludes $i$.  Smaller $u_i$ means that the prefix is less
isolated in the competitive pool.  We refer to this quantity as a local prefix
density proxy.  It is not a feasibility label and does not use executed
outcomes.

The proxy has a related finite-sample interpretation.  Define radius support
$S_r(i)$ as the fraction of other prefixes within radius $r$ of $\mathbf{p}_i$.
Suppose, conditional on the proposal source distributions, a feasible candidate
$e$ has $\mathbb{E}S_r(e)\geq p_+$ and an infeasible ranking blocker $b$ has
$\mathbb{E}S_r(b)\leq p_-$, with $p_+>p_-$.
For $n$ comparisons, Hoeffding's inequality~\citep{hoeffding1963probability} gives
\begin{equation}
    \Pr(S_r(e)\leq S_r(b))
    \leq 2\exp\!\left[-\frac{n(p_+-p_-)^2}{2}\right].
    \label{eq:concentration}
\end{equation}
The probability that the radius support statistic reverses this ordering is
therefore bounded under the stated support gap.  This result concerns a radius
count.  The implemented method uses KNN distance in Eq.~\ref{eq:support}. The radius
count and KNN distance both measure local prefix concentration, but they are
distinct statistics and are analyzed separately.

\subsection{Full Action Sequence Reconstruction}

\begin{figure*}[t]
\centering
\includegraphics[width=0.72\textwidth]{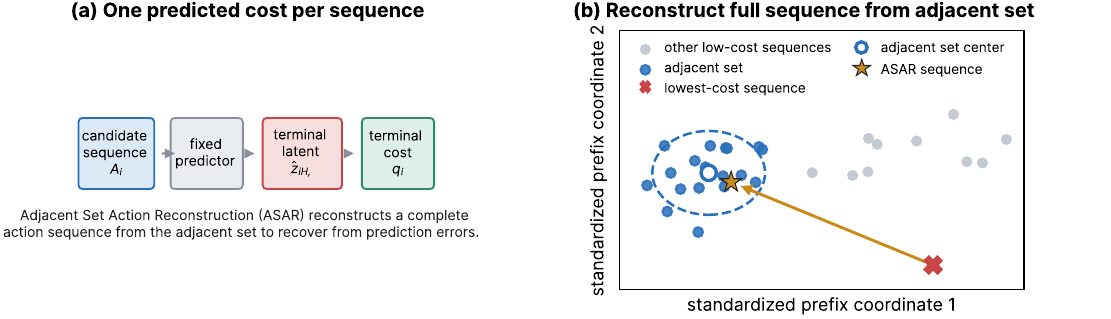}
\caption{Adjacent Set Action Reconstruction. \textbf{(a)} A fixed predictor maps each candidate action sequence to one terminal prediction and task cost. The cost may be latent distance or a trajectory reachability cost. \textbf{(b)} Among candidates with low cost, \asar{} uses standardized prefix distances to identify an adjacent set. It reconstructs a new full action sequence by weighted averaging and adds a 10\% anchor from the sequence with minimum cost. The plotted coordinates are a two-dimensional visualization of the prefix features, not the full action space.}
\label{fig:method}
\end{figure*}

Selecting the least isolated existing sequence uses the proxy but cannot produce
a sequence between sampled candidates.  Kernel \asar{} instead chooses the
candidate $c$ with smallest $u_i$, identifies an adjacent set $\mathcal{S}$ around its
prefix, and assigns
\begin{equation}
    w_i\propto
    \exp\!\left(-\frac{\|\mathbf{p}_i-\mathbf{p}_c\|_2^2}{\tau_d}
    -\lambda\tilde q_i\right),\quad i\in\mathcal{S},
\end{equation}
where $\tilde q_i$ is the robustly standardized terminal cost.  Let
$\A_Q=\arg\min_i q_i$ be the sequence with minimum cost in the eligible pool.  The
reconstruction operator produces a full action sequence
\begin{equation}
    \aRec=(1-\alpha)\sum_{i\in\mathcal{S}}w_i\A_i
    +\alpha\A_Q.
    \label{eq:asar_reconstruction}
\end{equation}
The first term is a weighted local sequence and the second is a light anchor from the sequence with minimum cost.  The anchor can help when the sequence with minimum cost is
already useful, while the local term reduces reliance on an isolated sequence.
The reconstruction is valid only when the action parameterization permits this
coordinatewise interpolation.  Figure~\ref{fig:method} shows the operation in
a two-dimensional visualization of prefix space.

The reconstructed sequence has a conditional containment property.  Suppose every full
sequence in $\mathcal{S}$ is within radius $R$ of a full-sequence center
$\A_c$ in a specified sequence norm and let
$D=\|\A_Q-\A_c\|$.
Nonnegative normalized weights give
\begin{equation}
    \|\aRec-\A_c\|
    \leq (1-\alpha)R+\alpha D.
    \label{eq:reconstruction_bound}
\end{equation}
If every full sequence in the ball
$\{\A:\|\A-\A_c\|\leq\gamma\}$ is feasible and the right-hand side is at
most $\gamma$, the reconstructed sequence is feasible.  The adjacent-set identification rule does
not by itself verify this full-sequence condition.  The supplement gives the
proof, exact parameter choices, a medoid variant, and the scalar support
re-ranking diagnostic.  The result is a local condition on the selected
sequence neighborhood and does not extend to arbitrary proposal pools.

\section{Experiments}

We ask three questions. First, does residual prediction error degrade the minimum cost selection as the proposal budget grows? Second, does full action sequence reconstruction improve over selecting an existing candidate using the same prefix density proxy? Third, do the gains persist across proposal budgets and independently trained world model weights, and where do they stop?

\paragraph{Environment and controller.}
We use simulated Cube manipulation from OGBench~\citep{park2024ogbench} with a
fixed pretrained \lewm{} encoder, predictor, and \cem{} generator.  The generator draws equally from action-noise scales 1.0, 1.5, and 2.0.  Total proposal budgets are 72, 144, and 288.  The controller has 40 environment steps.  Each model action represents a block of five environment actions, and the controller replans after each block. Kernel \asar{} retains the 80 lowest-cost sequences, computes $u_i$ from the first model action with $K=3$, takes the 12 nearest sequences around the least isolated center within the 32 least isolated candidates, and uses $\alpha=.10$.  These parameters are fixed across reported evaluations.  At evaluation time, no method observes the executed outcomes of candidates in the current pool.

\paragraph{Matched comparison of terminal costs.}
The Carry and Release evaluation set contains 75 queries defined by start and goal states from 44 evaluation seeds. Some seeds contribute more than one query, so confidence intervals resample seeds as clusters. Query membership is defined from the associated expert trajectory before planner outcomes are inspected. Contact is present at the start and absent at the goal, and the cube is lowered by more than 3 cm. We compare two terminal costs, latent distance and \trm{} reachability, with three ways to produce the next sequence. minimum cost selection and selection of the least isolated sequence each return an existing candidate. Kernel \asar{} reconstructs a new full action sequence. All six combinations use matched random proposal draws. The reachability head uses 100,000 trajectory pairs and excludes every episode represented in this evaluation set or the candidate execution scaling audit.

\paragraph{Outcomes and additional evaluation sets.}
The primary outcome for Carry and Release is event completion success. It requires the environment success indicator and a final contact state consistent with release. Other Cube evaluations report the environment success indicator. A rescue is a query where \asar{} succeeds and its stated baseline fails; a loss is the reverse. Confidence intervals are 95\% bootstrap intervals clustered by seed from 10,000 resamples.

A separate targeted test contains 73 queries from 54 new evaluation seeds. Its predefined rule uses translation between the start and goal, yaw change, and the maximum kernel weight in the initial pool. The rule selects cases where the reconstruction weight is spread across several adjacent candidates. This set is used to compare reconstruction with selection of the least isolated existing candidate, not to estimate performance over the full query distribution. We also report two Cube evaluations with 512 queries, including one with independently trained world model weights. Exact selection rules, episode exclusions, hyperparameter provenance, transfer tests, and all secondary controls appear in the supplement.

\section{Results}

\subsection{Adjacent-Set Reconstruction Improves under Both Terminal Costs}

\begin{figure*}[t]
\centering
\includegraphics[width=0.72\textwidth]{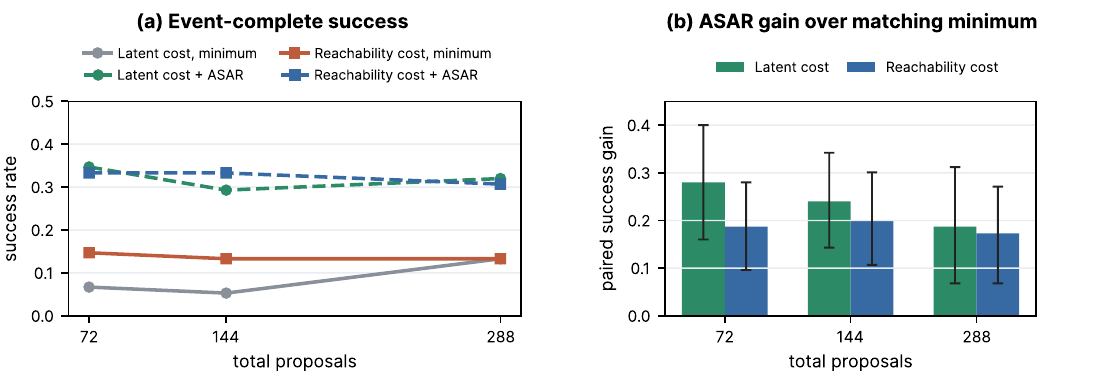}
\caption{Carry and Release results across total proposal budgets. (A) Event completion success for minimum cost selection and Kernel \asar{} under latent and reachability costs. (B) Paired Kernel \asar{} gain over the matching baseline based on minimum cost. Error bars are 95\% bootstrap intervals that resample the 44 evaluation seeds.}
\label{fig:factorial}
\end{figure*}

Table~\ref{tab:factorial} and Figure~\ref{fig:factorial} compare two terminal
costs and three output rules.  Under latent cost, Kernel \asar{} improves
event completion success over minimum cost selection by 28.0 percentage points
at 72 proposals, 24.0 points at 144, and 18.7 points at 288.  The corresponding
95\% intervals are $[16.0,40.0]$, $[14.3,34.2]$, and $[6.8,31.2]$ points.
Using the reachability cost gives gains of 18.7, 20.0, and 17.3 points with
intervals $[9.6,28.0]$, $[10.7,30.1]$, and $[6.8,27.1]$.

\begin{table*}[t]
\centering
\small
\setlength{\tabcolsep}{5pt}
\begin{tabular}{rrrrrrr}
\toprule
Total & \multicolumn{3}{c}{Latent cost} & \multicolumn{3}{c}{Reachability cost}\\
\cmidrule(lr){2-4}\cmidrule(lr){5-7}
proposals & Minimum & Least isolated & Kernel \asar{} & Minimum & Least isolated & Kernel \asar{}\\
\midrule
72  & .067 & .187 & \textbf{.347} & .147 & .187 & \textbf{.333}\\
144 & .053 & .213 & \textbf{.293} & .133 & .227 & \textbf{.333}\\
288 & .133 & .147 & \textbf{.320} & .133 & .213 & \textbf{.307}\\
\bottomrule
\end{tabular}
\caption{Event completion success on the same 75 Carry and Release queries from 44 evaluation seeds. Columns compare selection of an existing candidate with Kernel
\asar{} under two terminal costs. All methods use matched proposal draws.}
\label{tab:factorial}
\end{table*}

The baseline that selects the least isolated sequence controls for using the same prefix-density proxy
without reconstructing a new sequence.  Under reachability cost, Kernel \asar{}
exceeds this baseline by 14.7 points at 72 proposals with interval
$[3.9,25.0]$, and by 10.7 points at 144 with interval $[1.3,21.1]$.  At 288,
the gain is 9.3 points with interval $[-3.0,20.3]$.  Thus the first two budgets
give direct evidence for reconstruction beyond selection of an existing candidate,
while the larger-budget comparison remains uncertain.  The reachability cost
addresses terminal objective quality.  The consistent \asar{} gains over its
matching baseline based on minimum cost show that residual prediction failures remain a
separate control problem.

\subsection{Reconstruction Improves beyond Selection of an Existing Candidate}

The targeted 73-query test compares two rules built from the same 80 sequences with low cost and the same prefix distances.  Selecting the least isolated existing
sequence reaches .493 environment success.  Reconstructing a full action sequence with
Kernel \asar{} reaches .630, a gain of 13.7 percentage points with 95\%
interval $[1.9,23.8]$.  The corresponding mean errors in cube position, target position,
and yaw errors change from .154, .071, and .843 to .135, .047, and .762.
Table~\ref{tab:auxiliary} reports this comparison beside the aggregate and
evaluation with an independent model with each outcome and baseline stated explicitly.

\begin{table*}[t]
\centering
\scriptsize
\setlength{\tabcolsep}{3pt}
\begin{tabular}{llrlrrl}
\toprule
Evaluation set & Outcome & $N$ & Baseline rule & Baseline & Kernel \asar{} & $\Delta$ [95\% CI]\\
\midrule
Cube aggregate & Environment & 512 & Latent minimum & .723 & .762 & +.039 [+.016,+.065]\\
Independent model & Environment & 512 & Latent minimum & .703 & .738 & +.035 [+.014,+.057]\\
Targeted reconstruction & Environment & 73 & Least isolated & .493 & .630 & +.137 [+.019,+.238]\\
Carry-and-Release & Event complete & 75 & Latent minimum & .133 & .320 & +.187 [+.069,+.310]\\
\bottomrule
\end{tabular}
\caption{Closed-loop comparisons across Cube evaluation sets. Each row states its outcome and baseline, so rates and paired differences are interpreted only within that row. The row with an independent model uses 72 total proposals and the other rows use 288.}
\label{tab:auxiliary}
\end{table*}

With independently trained world model weights, the least isolated existing sequence reaches .717 and Kernel \asar{} reaches .738 at 72 proposals.  Their
paired difference is 2.1 points with interval $[1.0,3.5]$.  This directly
replicates the reconstruction advantage without relying on the original model
weights.  Figure~\ref{fig:carry_release} provides a physical example of the
same distinction between returning a sampled sequence and reconstructing a new
full action sequence.

\subsection{Where Adjacent-Set Reconstruction Helps}

On the 75 Carry and Release queries at 288 proposals, Kernel \asar{} rescues 19
queries that fail under selection by minimum latent cost and loses 5 queries that
the baseline solves.  The remaining 51 outcomes are unchanged.  A separate
512-query evaluation gives success under selection by minimum latent cost .711 and Kernel \asar{}
success .742, a paired gain of 3.1 points with interval $[1.0,5.5]$.  On the
42 queries whose expert segment contains grasp, lift, transport, lower, and
release within one interval from start to goal, however, success under selection by minimum latent cost is
.119 and Kernel \asar{} success is .048.  A single adjacent prefix set
does not supply the several distinct behaviors missing from these longer
queries.

Ordinary elite averaging reaches .721 on the same 512-query set.  Kernel
\asar{} is 2.15 points higher with interval $[0.0,4.3]$, and its continuous
position errors also improve with intervals above zero.  The supplement tests
other explanations using terminal state density, action index penalties,
filtering based only on the graph, and random filters matched to the graph retention rate.
None matches the Kernel point estimate on that evaluation set.

Portfolio \asar{} is also evaluated on two different
domains.  On Reacher, its success is .796 compared with .800 for the standard
\cem{} mean, with interval clustered by seed $[-.064,.044]$.  On Push-T, success under latent cost changes from .273 to .280 with interval $[-.080,.100]$; under a true
state hybrid cost it changes from .413 to .393.  These results show that the
Cube prefix metric is not a universal action-space metric.  The positive Cube
result appears when feasible sequences with low cost occupy a locally dense prefix
region that is suitable for interpolation.

\section{Discussion and Limitations}

Increasing a proposal budget changes both coverage and selection risk.  Added
candidates may improve \cem{} distribution estimation and expose new feasible
regions.  When predicted costs retain a lower tail for infeasible sequences,
minimum cost selection also receives more opportunities to choose one of those
errors.  Reporting feasible presence, top-$k$ feasible coverage, and mean
ranking-blocker count separates these effects more clearly than reporting
candidate count alone.

The cited \trm{} method and \asar{} act on different parts of the controller's decision rule. The former changes the terminal cost assigned to each predicted endpoint~\citep{li2026trm}, whereas \asar{} reconstructs the action sequence after the candidates have been scored. They can therefore be combined within the same controller.

Three conditions explain the scope of \asar{}. The terminal cost must define a useful candidate set with low costs. Feasible candidates should be less isolated in the chosen prefix metric than the ranking blockers. Reconstructed full sequences should remain inside a locally robust feasible neighborhood. If infeasible sequences form a dense cluster, prefix density favors the wrong region. If the only feasible sequence is isolated, density can reject it. If a query requires several distinct behaviors that are absent from one local neighborhood, averaging that neighborhood cannot create the missing sequence. A medoid avoids interpolation, graph filtering limits mixing across clusters, and a bounded controller mixture can test several fixed reconstruction variants in resettable settings. These variants address different failure cases rather than removing the three conditions.

The closed-loop evidence uses two independently trained models within one simulated Cube domain. Carry and Release begins from states in which the cube is already grasped, and the unchanged method is neutral on Reacher and neutral or negative on Push-T. The radius support result assumes a local probability gap. The containment result additionally assumes closeness of full sequences and a robust feasible ball, neither of which follows automatically from prefix KNN distance. Hardware evaluation should therefore measure sensing noise, failure recovery, and the cost of executing alternative controllers from a safe reset state.

\section{Conclusion}

A controller that selects the minimum predicted terminal cost can become less reliable when a larger proposal pool contains more prediction errors with low cost.
\asar{} identifies an adjacent set among action prefixes with low cost and reconstructs a weighted full action sequence, then executes its first action block in the usual receding-horizon loop. Candidate execution audits identify the ranking failure, matched closed-loop comparisons show gains under two terminal costs, and the targeted reconstruction test separates the new sequence from selection of an existing dense candidate. Proposal sets are useful not only for finding a low predicted cost, but also for estimating whether that candidate belongs to an adjacent set of other plausible actions.

\clearpage
\appendix

\section{Appendix Overview}

This supplement expands the diagnosis, method, theory, and experimental protocol in the main paper. It first defines the physical execution setting and specifies each \asar{} reconstruction variant. It then states the finite proposal pool results, records the Cube protocol, and reports candidate execution analyses and matched closed-loop comparisons. Later sections provide replication with independently trained world model weights, joint latent and action neighborhood controls, a targeted reconstruction evaluation, a bounded mixture of controllers, and transfer results on Reacher and Push-T.

Cube is the main method setting because its low cost proposal pools contain both ranking blockers and locally dense action prefixes. Reacher and Push-T test the unchanged Cube reconstruction rule when this prefix geometry differs. The theory mirrors this scope. The lower tail identity concerns costs in a finite pool. The radius support result requires an explicit local probability gap. The reconstruction bound additionally requires closeness of the selected full sequences and a robust feasible neighborhood around their center. Physical value is measured through paired closed-loop execution.

\section{Notation and Physical Execution}

Let \(\x_t\in\mathcal{X}\) be the physical state at one replanning step and
let \(o_t=g(\x_t)\) be the corresponding observation.  The encoder \(e\)
produces \(\z_t=e(o_t)\) and the goal latent \(\zg=e(o_g)\).  A candidate
action sequence is
\begin{equation}
    \A_i=(\mathbf{a}_{i,0},\ldots,\mathbf{a}_{i,H-1})\in\mathcal{A}^{H}.
\end{equation}
One model action \(\mathbf{a}_{i,h}\) is an action block that represents five
environment actions in Cube.  An action prefix contains the first \(h\) model
actions of the full sequence.
The learned predictor maps the current latent and action sequence to a
predicted terminal latent,
\begin{equation}
    \hat{\z}_{i,H}=\hat f_H(\z_t,\A_i).
    \label{eq:supp_predicted_latent}
\end{equation}

In a candidate execution audit, \(F_H\) denotes counterfactual physical
execution of the full sequence.  The terminal state and its encoded observation
are
\begin{equation}
    \x^{\mathrm{exec}}_{i,H}=F_H(\x_t,\A_i),\qquad
    \z^{\mathrm{exec}}_{i,H}=e\!\left(g(\x^{\mathrm{exec}}_{i,H})\right).
    \label{eq:supp_executed_latent}
\end{equation}
Equations~\ref{eq:supp_predicted_latent} and
\ref{eq:supp_executed_latent} name different objects.  The planner ranks
\(\hat{\z}_{i,H}\).  The audit obtains
\(\z^{\mathrm{exec}}_{i,H}\) only after executing candidate \(i\) from the
same stored initial state.  Closed-loop control does not execute \(F_H\) for
every candidate.  It executes only the first action block of the selected
sequence, observes the next image, and replans.

A terminal cost maps the predicted endpoint and goal to one scalar,
\begin{equation}
    q_i=Q(\hat{\z}_{i,H},\zg).
    \label{eq:supp_supplied_score}
\end{equation}
Lower values are preferred.  Latent distance and \trm{} reachability are the
two evaluated instances of \(Q\).  All subsequent prefix operations condition
on this pool ordered by terminal cost.

Let \(\ell_H\) include the full horizon terminal conditions used by an audit,
including target error and release state when required.  For tolerance
\(\epsilon\), the feasible sequence set is
\begin{equation}
    \mathcal{E}_{\epsilon}(\x_t,o_g)
    =\{\A:\ell_H(F_H(\x_t,\A),o_g)\leq\epsilon\}.
    \label{eq:supp_feasible_set}
\end{equation}
Inside a finite proposal pool \(\mathcal{C}_B\), let \(Y_i=1\) indicate
\(\A_i\in\mathcal{E}_{\epsilon}\).  A ranking blocker under \(Q\) is an
infeasible proposal \(b\) whose cost is lower than every feasible cost,
\begin{equation}
    Y_b=0,\qquad
    q_b<\min_{i:Y_i=1}q_i.
    \label{eq:supp_blocker}
\end{equation}
The total proposal budget \(B\) counts action sequences in one replanning pool.
A query denotes one closed-loop evaluation from a start state to a goal state and may contain several
replanning pools.

\section{Detailed ASAR Implementations}

\subsection{Candidate Pool and Terminal Cost}

At each replan, the fixed world model receives the current image, the goal
image, and a shared pool of action sequences
\(\{\A_i\}_{i=1}^{N}\), where
\(\A_i\in\R^{H\times d_a}\).  It predicts terminal latent states
\(\hat{\z}_i\).  The terminal cost rule assigns
\begin{equation}
    q_i=Q(\hat{\z}_i,\zg).
    \label{eq:supp_score}
\end{equation}
For latent cost, \(q_i=\|\hat{\z}_i-\zg\|_2^2\).  For reachability cost,
\(Q\) is the learned temporal head described in
Section~\ref{sec:factorial_protocol}.  All output rules compared under one cost
use the same initial state, goal, fixed world model, proposal generator,
matched proposal noise, and execution budget.

For the Cube evaluations, the proposal generator produces 96 sequences at
each of three variance scales \(1.0,1.5,2.0\), for 288 candidates per replan.
The \asar{} reconstruction variants first retain the 80 candidates with lowest terminal cost.
This restriction preserves goal competitiveness while
excluding obviously poor proposals.

\subsection{Local Prefix Isolation Score}

Let \(\mathbf{p}_i\) be the vectorized first model action of candidate \(i\).
For Cube this prefix contains one action block, corresponding
to five environment actions.  Each prefix coordinate is robustly standardized
within the current eligible pool using the median and interquartile range,
falling back to the standard deviation when the interquartile range is
degenerate.

For each eligible candidate, the implementation computes the mean Euclidean
distance to its \(k=3\) nearest other standardized prefixes,
\begin{equation}
    u_i=\frac{1}{k}\sum_{j\in\mathcal{N}_k(i)}
    \|\mathbf{p}_i-\mathbf{p}_j\|_2.
    \label{eq:supp_knn}
\end{equation}
Smaller distance means that a prefix is less isolated in the eligible pool.
The distances are robustly standardized again before use.  Sorting by \(u_i\)
defines the prefix density order.  The least isolated baseline returns the
existing candidate at the head of this order.  The two main \asar{} variants
instead reconstruct a full action sequence from an adjacent set.  This statistic is available at planning time and serves as a density proxy, but it does not by itself establish
physical feasibility.

\subsection{Kernel \asar{} Sequence Reconstruction}

The released evaluator implements Kernel \asar{} with the fixed parameters summarized in Table~\ref{tab:kernel_params}. The terminal cost term in the kernel weight uses the robustly standardized cost defined for the current proposal pool.

\begin{table}[tbp]
\centering
\small
\begin{tabular}{lr}
\toprule
Item & Value\\
\midrule
Eligible low cost candidates & 80\\
Prefix steps & 1\\
KNN neighbors & 3\\
Prefix density candidate pool & 32\\
Adjacent set size & 12\\
Kernel temperature \(\tau\) & 0.75\\
Terminal cost energy weight & 0.35\\
Minimum cost anchor \(\alpha\) & 0.10\\
\bottomrule
\end{tabular}
\caption{Fixed Kernel \asar{} parameters used in the confirmatory runs.}
\label{tab:kernel_params}
\end{table}

The least isolated candidate supplies the prefix center.  Among the first 32
candidates in prefix density order, the method identifies the 12 prefixes nearest
to this center as the adjacent set.  Let \(D\) be the prefix feature
dimension, let \(\bar d_i^2=\|\mathbf{p}_i-\mathbf{p}_c\|_2^2/D\), and let
\(\tilde q_i\) denote the robustly standardized terminal cost.  The
unnormalized weight
is
\begin{equation}
    \tilde w_i =
    \exp\left[-\left(\frac{\bar d_i^2}{0.75}
    +0.35\tilde q_i\right)\right].
    \label{eq:kernel_weight}
\end{equation}
After normalization, the sequence reconstructed from the adjacent set is
\begin{equation}
    \A_{\mathrm{ker}}=\sum_{i\in\mathcal{S}_{12}} w_i\A_i.
\end{equation}
The output retains the sequence with minimum cost \(\A_Q\):
\begin{equation}
    \A_{\mathrm{out}}=0.90\A_{\mathrm{ker}}+0.10\A_Q.
    \label{eq:kernel_anchor}
\end{equation}
No executed outcome, success label, or physical state obtained from the simulator enters the reconstruction operator.

\subsection{Portfolio \asar{} Sequence Reconstruction}

Portfolio \asar{} identifies the seven least isolated candidates as an adjacent set
\(\mathcal{E}\).  In robustly standardized prefix space, define
\begin{equation}
    \rho =
    \frac{\|\mathbf{p}_{Q}-\bar{\mathbf{p}}_{\mathcal{E}}\|_2}
    {\frac{1}{|\mathcal{E}|}
    \sum_{i\in\mathcal{E}}
    \|\mathbf{p}_{i}-\bar{\mathbf{p}}_{\mathcal{E}}\|_2+\epsilon}.
    \label{eq:portfolio_ratio}
\end{equation}
The fixed branch rule is:
\begin{itemize}
    \item \textbf{Close, \(\rho\leq1.0\):} use the elite arithmetic mean and
    anchor from the sequence with minimum cost \(\alpha=0.25\).
    \item \textbf{Middle, \(1.0<\rho<1.75\):} use the elite medoid in
    prefix space and anchor from the sequence with minimum cost \(\alpha=0.15\).
    \item \textbf{Far, \(\rho\geq1.75\):} use the geometric median of the
    full elite action sequences and anchor from the sequence with minimum cost \(\alpha=0.10\).
\end{itemize}
For representative \(\A_{\mathrm{rep}}\), the output is always
\begin{equation}
    \A_{\mathrm{out}}=(1-\alpha)\A_{\mathrm{rep}}
    +\alpha\A_Q.
    \label{eq:portfolio_anchor}
\end{equation}
The medoid branch outputs an original sampled candidate. The mean and geometric median branches reconstruct a sequence from the adjacent set. Proposition 5 states the additional local conditions under which the reconstructed sequence remains inside the feasible set.

\subsection{Joint Latent and Action Neighborhood Reconstruction}

The joint neighborhood extension first filters the 80 lowest cost candidates before applying the same Portfolio ASAR rule. For each candidate, robustly standardized predicted terminal latent features are concatenated with the robustly standardized action prefix containing one model action. The joint features are reduced to at most 16 principal components. An undirected graph is then formed by connecting each candidate to its eight nearest neighbors.

The five nodes with the smallest mean neighbor distance define the dense neighborhood seeds. Shortest path distance from these seeds is computed for all eligible candidates. The method retains the lower 50\% of these distances, subject to a minimum of 16 and a maximum of 48 candidates. With 80 eligible candidates, the retained neighborhood contains 40 candidates. The Portfolio ASAR prefix density order, elite set, branch ratio, representative, and minimum cost anchor are recomputed inside this retained neighborhood. The anchor is the candidate with minimum cost inside the retained neighborhood, not the global cost minimum.

The random control matched to the retention rate uses the same retained size and is calibrated to match the joint neighborhood method's empirical rate of filtering out the candidate with minimum cost. It therefore tests whether the result follows from generic pruning rather than filtering based on the joint latent and action neighborhood.

\subsection{Optional Bounded Mixture of Controllers}

The Mixture of Controllers (\moc{}) is a supplementary bounded verification extension rather than part of the single execution \asar{} method. It treats uncertainty across reconstruction variants as a bounded verification problem. For each query it executes a fixed controller order: Portfolio \asar{}, Kernel \asar{}, anchored adjacent prefix mean, and adjacent prefix mean. If an earlier branch succeeds within the 40-step Cube budget, evaluation stops. If it fails, the next branch is run from the same initial state and goal. Thus, the protocol allocates a bounded branch budget across controller families and evaluates them sequentially, stopping at the first successful branch.

The four concrete output rules are:
\begin{enumerate}
    \item Portfolio \asar{};
    \item Kernel \asar{};
    \item the adjacent prefix mean with a 0.25 anchor on the sequence with minimum cost;
    \item the unanchored adjacent prefix mean.
\end{enumerate}
The verifier never inspects later outcomes before deciding to continue. In the current protocol, failed branches reset to the same initial state. This matches repeatable simulation and resettable control settings. Irreversible deployment requires a safe state recovery mechanism in place of reset.

\subsection{Planning Procedure}

At each replan, \asar{} performs the following operations using information available at planning time:
\begin{enumerate}
    \item Sample candidates at the fixed proposal scales and roll them through the fixed world model.
    \item Compute the terminal cost and retain its 80 lowest cost candidates.
    \item Extract the first action block, robustly standardize the prefixes,
    and compute the \(k=3\) nearest-neighbor prefix isolation score.
    \item For the kernel variant, restrict to the 32 least isolated
    candidates, take the nearest 12, and apply
    Eqs.~\ref{eq:kernel_weight}--\ref{eq:kernel_anchor}.  For the portfolio,
    take seven adjacent prefix candidates and apply
    Eqs.~\ref{eq:portfolio_ratio}--\ref{eq:portfolio_anchor}.  The graph
    extension first filters to the retained joint latent and action neighborhood and then applies the same portfolio rule.
    \item Execute the first five environment actions, observe the next image, and repeat until success or the 40-step Cube budget is exhausted.
\end{enumerate}
All operations use only the current proposal pool, predicted latent rollouts, terminal costs, and action sequences.

\moc{} wraps four complete instances of this planner. It evaluates them in the fixed order given above, stops on the environment success signal, and resets to the same stored initial state only after a failed branch. The controller order and branch limit do not depend on outcomes from later controllers.

\section{Finite Proposal Pool Results and Proofs}

The controller chooses from a finite proposal pool at each replan. The statements below analyze that decision object directly using proposal costs, adjacent prefix sets, and reconstructed action sequences. This pool level view also accommodates contact transitions where the physical map is not globally continuous.

\subsection{Growth of Lower Tail Ranking Errors}

\paragraph{Proposition 1 (risk from lower tail overgeneration).}
Condition on \(n_-\) infeasible proposals.  Suppose their terminal costs are
conditionally independent with lower tail distribution
\(F_-(t)=\Pr(q_b\leq t\mid Y_b=0)\).  Then
\begin{equation}
    \Pr\!\left(\min_{b:Y_b=0}q_b\leq t\right)
    =1-\left(1-F_-(t)\right)^{n_-}.
    \label{eq:supp_lower_tail}
\end{equation}
For any threshold with \(0<F_-(t)<1\), the probability is strictly increasing
in \(n_-\).

\paragraph{Proof.}
The event \(\min_b q_b>t\) requires every infeasible proposal to have score
greater than \(t\).  Conditional independence gives
\(\Pr(\min_b q_b>t)=(1-F_-(t))^{n_-}\).  Taking the complement proves
Eq.~\ref{eq:supp_lower_tail}.  Its one step increase is
\(F_-(t)(1-F_-(t))^{n_-}>0\). \(\square\)

The independence model makes the lower tail mechanism explicit.  The expected
blocker count does not need this assumption.  Condition on the best feasible
score \(T_B\) and define \(I_j=\mathbf{1}[q_j<T_B]\) for each infeasible
proposal.  Linearity of expectation gives
\begin{equation}
    \mathbb{E}[M_B\mid T_B]
    =\sum_{j=1}^{N_B}\Pr(q_j<T_B\mid T_B).
    \label{eq:supp_blocker_expectation}
\end{equation}
Under conditionally identical tails, this becomes \(N_BF_-(T_B)\).  Correlation
changes the count distribution but not the interpretation that a larger pool
exposes more lower tail opportunities.

\paragraph{Proposition 2 (presence and top-\(k\) retention).}
At budget \(B\), let \(P_B\) be the event that the proposal pool contains at
least one feasible sequence.  Let \(T_{B,k}\) be the event that at least one
feasible sequence appears among the \(k\) lowest cost proposals.  Since
\(T_{B,k}\subseteq P_B\),
\begin{equation}
    \Pr(T_{B,k})=\Pr(P_B)\Pr(T_{B,k}\mid P_B).
    \label{eq:supp_decomp}
\end{equation}

\paragraph{Proof.}
The set identity \(T_{B,k}=T_{B,k}\cap P_B\) and the multiplication rule give
Eq.~\ref{eq:supp_decomp}. \(\square\)

The two factors distinguish candidate generation from retention based on cost.
When \(\Pr(P_B)\) is already high, a reduction in
\(\Pr(T_{B,k}\mid P_B)\) can lower the probability that a feasible sequence is
available to a top-\(k\) output rule.

\subsection{Concentration of Local Prefix Support}

For a radius \(r\), define the support of proposal \(i\) among \(n\) other
proposals by
\begin{equation}
    S_r(i)=\frac{1}{n}\sum_{j=1}^{n}
    \mathbf{1}[\rho(\mathbf{p}_i,\mathbf{p}_j)\leq r].
    \label{eq:supp_radius_support}
\end{equation}

\paragraph{Proposition 3 (local support separation).}
Let \(e\) be a feasible candidate and let \(b\) be a ranking blocker.  Suppose
the indicators of neighbors within radius $r$ are conditionally independent for each candidate
and satisfy
\begin{equation}
    \mathbb{E}[S_r(e)]\geq p_+,
    \qquad
    \mathbb{E}[S_r(b)]\leq p_-,
    \qquad p_+>p_-.
\end{equation}
Then
\begin{equation}
    \Pr(S_r(e)\leq S_r(b))
    \leq 2\exp\!\left[-\frac{n(p_+-p_-)^2}{2}\right].
    \label{eq:supp_concentration}
\end{equation}

\paragraph{Proof.}
Let \(m=(p_++p_-)/2\) and \(\Delta=p_+-p_-\).  If
\(S_r(e)\leq S_r(b)\), then either \(S_r(e)\leq m\) or
\(S_r(b)\geq m\).  Hoeffding's inequality gives
\begin{align}
    \Pr(S_r(e)\leq m)&\leq\exp(-n\Delta^2/2),\\
    \Pr(S_r(b)\geq m)&\leq\exp(-n\Delta^2/2).
\end{align}
The union bound proves Eq.~\ref{eq:supp_concentration}. \(\square\)

The radius count makes the finite sample argument transparent.  The implemented
statistic in Eq.~\ref{eq:supp_knn} instead measures the distance needed to reach
\(k\) neighbors.  Both summarize local prefix density, with opposite numerical
directions: larger \(S_r\) and smaller \(u_i\) indicate a less isolated prefix.
The candidate execution audit in Section~\ref{sec:gap_audit} evaluates the exact
KNN statistic used by the controller.

\subsection{Existing Candidate Reranking by Prefix Isolation}

\paragraph{Proposition 4 (condition for pairwise reranking).}
Let \(e\) be a feasible candidate and \(b\) a ranking blocker with
\(q_b<q_e\).  Let \(u_i\) be the observable prefix isolation score from
Eq.~\ref{eq:supp_knn}, and define \(s_i=q_i+\beta u_i\) with
\(\beta>0\).  Then
\begin{equation}
    u_b-u_e > \frac{q_e-q_b}{\beta}
    \label{eq:supp_gap}
\end{equation}
if and only if \(s_e<s_b\).

\paragraph{Proof.}
Multiplying Eq.~\ref{eq:supp_gap} by \(\beta\) and rearranging gives
\(q_e+\beta u_e<q_b+\beta u_b\).  This is exactly \(s_e<s_b\).  Reversing
the algebra proves the converse. \(\square\)

Proposition 4 describes a rule for reranking existing candidates.  The audit in
Section~\ref{sec:gap_audit} measures its premise with executed candidates.
Kernel \asar{} and Portfolio \asar{} use the same local geometry to reconstruct a
new full action sequence, whose outcome is measured through paired closed-loop
execution.

\subsection{Containment of Reconstructed Full Sequences}

\paragraph{Proposition 5 (containment of a reconstructed sequence).}
Let \(\mathcal{S}\) be an adjacent set identified by prefix geometry, and let
\(\A_c\) be a reference full sequence.  Suppose
\(\|\A_i-\A_c\|\leq R\) for every \(i\in\mathcal{S}\), the weights satisfy
\(w_i\geq0\) and \(\sum_i w_i=1\), and the sequence with minimum cost obeys
\(\|\A_Q-\A_c\|=D\).  For
\begin{equation}
    \A_{\mathrm{out}}
    =(1-\alpha)\sum_{i\in\mathcal{S}}w_i\A_i+\alpha\A_Q,
\end{equation}
we have
\begin{equation}
    \|\A_{\mathrm{out}}-\A_c\|
    \leq (1-\alpha)R+\alpha D.
    \label{eq:supp_reconstruction_bound}
\end{equation}
If the closed ball \(\{\A:\|\A-\A_c\|\leq\gamma\}\) lies inside
\(\mathcal{E}_\epsilon\) and the right-hand side of
Eq.~\ref{eq:supp_reconstruction_bound} is at most \(\gamma\), then
\(\A_{\mathrm{out}}\in\mathcal{E}_\epsilon\).

\paragraph{Proof.}
The triangle inequality and convexity of the norm give
\begin{align}
\|\A_{\mathrm{out}}-\A_c\|
&\leq (1-\alpha)
\left\|\sum_iw_i(\A_i-\A_c)\right\|\notag\\
&\quad+\alpha\|\A_Q-\A_c\|\\
&\leq (1-\alpha)\sum_iw_i\|\A_i-\A_c\|+\alpha D\\
&\leq (1-\alpha)R+\alpha D.
\end{align}
The final statement follows from containment in the robust feasible ball.
\(\square\)

Proposition 5 formalizes the two reconstruction terms.  A compact reconstruction
set reduces \(R\), while the anchor from the sequence with minimum cost contributes according to its
displacement \(D\).  Here \(R\), \(D\), and \(\gamma\) are measured in the
space of complete action sequences.  Prefix proximity selects the reconstruction
set but does not by itself imply this full sequence bound.  The bound therefore
states the additional local condition under which weighted reconstruction remains
inside a robust feasible neighborhood.

\subsection{Separation of Proposal Neighborhoods}

\paragraph{Proposition 6 (separation of proposal clusters).}
Suppose a proposal pool is partitioned into clusters and every pair from different clusters is separated by more than \(\delta\) in the graph metric.  If graph edges
connect only pairs at distance at most \(\epsilon<\delta\), then no connected
component contains proposals from two clusters.

\paragraph{Proof.}
An edge between different clusters would require a pair from different clusters at
distance at most \(\epsilon\), contradicting the assumed separation.  Any path
that crosses clusters must contain such an edge. \(\square\)

The medoid of a component is one of its sampled proposals. Selecting that medoid therefore introduces no interpolation beyond the original candidate set. Proposition 5 gives the corresponding local condition for a weighted mean with a score anchor. The medoid branch remains available when reconstruction outside the sampled set is undesirable. Proposition 6 concerns separation in the observed proposal graph and does not assign a physical regime label to a component.

\subsection{Optional Bounded Mixture of Controllers}

\paragraph{Proposition 7 (success union and expected branch count).}
Let \(S_j\) be the success event for controller \(j\) from the shared initial
query, and let \moc{} execute controllers in a fixed order, stopping at the
first verified success.  Then
\begin{equation}
    \Pr(\moc\text{ succeeds})
    =\Pr\!\left(\bigcup_{j=1}^{J}S_j\right).
    \label{eq:supp_moc_union}
\end{equation}
If \(L\) is the number of executed branches, then
\begin{equation}
    \mathbb{E}[L]
    =\sum_{j=1}^{J}\Pr\!\left(\bigcap_{\ell<j}S_\ell^c\right).
    \label{eq:supp_moc_cost}
\end{equation}

\paragraph{Proof.}
The verifier returns success exactly when at least one branch succeeds, which
proves Eq.~\ref{eq:supp_moc_union}.  Branch \(j\) is executed exactly when all
earlier branches fail.  Since
\(L=\sum_{j=1}^{J}\mathbf{1}[L\geq j]\), linearity of expectation gives
Eq.~\ref{eq:supp_moc_cost}. \(\square\)

The result requires no independence among controller outcomes. It states the benefit and branch cost measured by validation from the same initial state on 768 new queries.

\FloatBarrier
\section{Cube Evaluation Protocol}

This section fixes the matched closed-loop protocol used for every Cube method comparison and records the disjoint evaluation splits and reproducibility information. ASAR receives only quantities visible to the planner; executed candidate outcomes are reserved for evaluation and the separate diagnostic audits.

\subsection{Closed-Loop Evaluation Configuration}

\begin{table*}[tbp]
\centering
\small
\begin{tabular}{L{0.25\textwidth}L{0.66\textwidth}}
\toprule
Item & Fixed setting\\
\midrule
Environment and data & \texttt{ogbench/cube\_single\_expert}~\citep{park2024ogbench};
fixed pretrained \texttt{cube/lewm} encoder and dynamics.\\
Queries & 8 paired queries per seed; identical start and goal for every method
within a query.\\
Goal offset and budget & Goal offset 40 environment steps; evaluation budget
40 environment steps.\\
Model horizon & 8 model actions, each representing an action block of 5
environment actions.\\
Replanning & Execute 5 environment actions, then replan; at most 8 replans.\\
Proposal pool & 96 candidates at each variance scale \(1.0,1.5,2.0\), giving
288 candidates per replan.\\
Eligible sequences & 80 lowest terminal costs; prefix length 1; \(k=3\)
nearest-neighbor isolation score.\\
Execution & Every reconstructed sequence is executed in the same environment
loop as the minimum cost sequence.  No oracle that executes every candidate is used by the
method.\\
Statistics & Paired success wins/losses; primary intervals resample evaluation
seeds as clusters with 10,000 bootstrap repetitions.  Intervals paired by query
are retained as sensitivity analyses.  Continuous improvements are baseline
minus method so positive is better.\\
\bottomrule
\end{tabular}
\caption{Cube closed-loop evaluation protocol.}
\label{tab:cube_protocol}
\end{table*}

\subsection{Data Splits and Evaluation Roles}

\begin{table*}[tbp]
\centering
\small
\begin{tabular}{L{0.27\textwidth}L{0.18\textwidth}L{0.20\textwidth}L{0.25\textwidth}}
\toprule
Experiment family & Scale & Role & Separation rule\\
\midrule
Proposal overgeneration & 12 development blocks, 4 held-out blocks, and 4 replication blocks & Diagnosis and replication & Held-out and replication blocks are not used for development\\
Canonical Cube evaluation & Two sets of 256 queries & Confirmatory method evaluation & Evaluation seeds are disjoint from method development\\
Final proposal pool controls & 512 queries & Output rule controls & Seven output rules are fixed before execution\\
Joint neighborhood study & 256 development and 256 control queries & Extension and matched control & The control evaluation is separate from development\\
Prefix density analysis & 96 calibration and 96 held-out pools & Analysis of the candidate level premise & The calibration coefficient is fixed before the held-out analysis\\
Independent model replication & Two sets of 512 queries & Replication across model weights & Uses an independently trained Cube world model\\
Transfer evaluations & 250 Reacher episodes and 150 paired Push-T episodes for each terminal cost & Boundary analysis & Uses the Cube reconstruction rule without parameter changes\\
Mixture of Controllers & 768 queries & Supplementary evaluation with resets & Controller order is fixed before evaluation\\
\bottomrule
\end{tabular}
\caption{Experimental roles and split boundaries. Exact seed lists, file names, and result mappings are recorded in the accompanying Code and Data Supplement.}
\label{tab:provenance}
\end{table*}

\subsection{Method Development and Confirmatory Evaluation Protocol}

The eligible set size, prefix length, neighbor count, kernel parameters, and Portfolio \asar{} branch settings were selected on development blocks before the canonical Cube evaluations. The joint neighborhood rule was developed on one query block and then applied without modification to a separate control block. The seven final proposal pool output rules were recorded before their evaluation. Exact seed assignments and configuration records are included in the accompanying artifact.

The independent model replication fixes the independently trained Cube checkpoint, proposal budgets, output rules, and inclusion rule before the replication evaluation. The Cube reconstruction rule is unchanged. The \moc{} controller order was recorded after the router analysis and before its separate evaluation. Its primary comparison is \moc{} success against Portfolio first execution under repeated attempts from the same initial state.

Table~\ref{tab:development_choices} summarizes the comparisons used to select the final reconstruction variants. The anchor study evaluated weights \(0,.10,.25,.35,.50\). It identified a small anchor weight on the sequence with minimum cost as the useful regime and showed that weights .35 and .50 attenuate the prefix density effect. The remaining parameters were fixed before the confirmatory evaluations.

\begin{table*}[tbp]
\centering
\small
\begin{tabular}{L{0.22\textwidth}L{0.40\textwidth}L{0.27\textwidth}}
\toprule
Component & Development comparison & Confirmatory setting\\
\midrule
Eligible set and local geometry & Competitiveness under minimum cost, prefix density, terminal density, and full sequence controls & 80 lowest costs, prefix 1, \(k=3\)\\
Minimum cost anchor & \(0,.10,.25,.35,.50\) on development pools & .10 for Kernel and .25/.15/.10 by Portfolio branch\\
Portfolio representative & Local means, medoids, robust centers, and elite size variants & Seven elites with ratio thresholds 1.0 and 1.75\\
Kernel ASAR & Local kernel centers, adjacent set sizes, terminal cost weights, and anchors & Pool 32, set 12, \(\tau=.75\), cost weight .35, anchor .10\\
Graph neighborhood & Latent, action, and joint graphs with random controls matched to the filtering rate & PCA 16, graph \(k=8\), five core seeds, median core distance retention\\
\bottomrule
\end{tabular}
\caption{Development choices and settings used without change in the confirmatory evaluations.}
\label{tab:development_choices}
\end{table*}

\section{Matched Evaluation of Terminal Costs and Output Rules}
\label{sec:factorial_protocol}

\paragraph{Relation to the trajectory reachability cost.}
Li et al. introduced Trajectory Reachability Metrics (\trm{}) to test whether terminal latent distance reflects physical task progress when scoring a predicted endpoint~\citep{li2026trm}. The learned metric changes the terminal cost assigned to existing candidates. \asar{} operates after this scoring step: it uses the geometry of early action prefixes in the current proposal pool and may reconstruct a full sequence that was not sampled. The matched comparison therefore crosses latent distance and \trm{} reachability with three output rules while holding the world model, proposal generator, evaluation queries, and proposal draws fixed.

\paragraph{Evaluation set.}
The set contains 75 Carry and Release queries grouped into 44 evaluation seeds. Query membership was recorded before the matched comparison using the same rule, defined independently of outcomes, as in the main paper. The associated expert segment begins in contact, ends without contact, and lowers the cube by more than 3 cm. The query definition file records the dataset row, episode, start and goal steps, evaluation seed, and query identifier for every case.

\paragraph{Episode exclusion.}
The temporal head is trained on the public
\path{ogbench/cube_single_expert} trajectories~\citep{park2024ogbench} after
excluding 102 episodes.
This union contains all 73 episodes represented by the Carry and Release
evaluation set and all 32 episodes in the independent candidate execution
set.
Three episodes occur in both sets.  The SHA-256 of the exclusion list is
\texttt{60ff037aed46ab53e6e1d33c34da14e2c}\allowbreak
\texttt{97fd37ebfee6926f60e791cbe7d67f4}.

\paragraph{Temporally Supervised Heads and Heads with Shuffled Labels.}
Both heads use the same pairwise architecture, 100,000 training pairs, 10,000 validation pairs, temporal range 1 to 40, target scale 40, optimizer schedule, and pair split. The temporal head learns within episode time separation. The control permutes only the training labels. The best validation RMSE is 8.97 for temporal supervision and 11.62 for shuffled supervision. Head training encodes 125,785 unique dataset rows and uses no evaluation episode.

\paragraph{Six matched controllers.}
For each terminal cost, latent distance or \trm{} reachability, the controller
uses one of three output rules.  Selection by minimum cost returns the existing
candidate with minimum \(q_i\).  Selection of the least isolated candidate returns the existing
candidate with the smallest prefix isolation score among the 80 lowest cost
candidates.  Kernel \asar{} uses the same eligible set and prefix geometry, then
reconstructs the anchored full sequence in Eq.~\ref{eq:kernel_anchor}.  Proposal
budgets are 24, 48, and 96 per source, or 72, 144, and 288 in total.  The six
controllers use matched evaluation seeds and proposal noise in one batched
environment call.

\paragraph{Analysis.}
Event completion success requires the environment success indicator and final contact error at most 0.1. Contrasts are paired at the query level. Confidence intervals resample the 44 seeds as clusters with replacement for 10,000 bootstrap draws. The main contrasts compare Kernel \asar{} with minimum cost selection and with least isolated existing sequence selection under each terminal cost. The difference between reconstruction gains estimates the interaction between terminal cost and output rule. The shuffled head run is evaluated at the largest proposal budget.

\section{Complete Proposal Overgeneration Results}

\subsection{Candidate Execution Diagnosis}

The first audit executes candidates from 192 fixed proposal pools. It records whether the latent cost minimum succeeds and whether any successful candidate is already present among the 80 lowest latent costs. The second diagnostic uses 48 separate proposal pools and compares selection under predicted terminal latent distance, executed terminal latent distance, and executed terminal task loss.

\begin{table}[!ht]
\centering
\small
\setlength{\tabcolsep}{4pt}
\resizebox{\columnwidth}{!}{%
\begin{tabular}{lrr}
\toprule
192-pool accounting & Pools & Rate\\
\midrule
Latent minimum succeeds & 60 & .313\\
Latent minimum fails, top-80 success exists & 58 & .302\\
Latent minimum fails, no top-80 success & 74 & .385\\
\midrule
48-pool selection cost & Success & \(\Delta\) predicted\\
\midrule
Predicted terminal latent & .292 & --\\
Executed terminal latent & .500 & +.208\\
Physical terminal loss & .542 & +.250\\
\bottomrule
\end{tabular}}
\caption{Candidate execution diagnosis. The 192-pool accounting and 48-pool terminal state comparison use separate fixed candidate sets.}
\label{tab:executed_pool_diagnosis}
\end{table}

Among the 132 pools where selection by minimum latent cost fails, 58 contain a successful candidate among the 80 lowest latent costs. This is 43.9\% of failures under selection by minimum latent cost. In these pools the candidate has been generated, but predicted latent cost ranks it behind a failed candidate. The separate 48-pool result shows that this order changes substantially when the terminal latent is encoded after execution.

\subsection{Replication across Proposal Budgets}

Table~\ref{tab:overgen_all} reports the independent replication.  The
position and yaw targets reach complete feasible presence at 24
candidates per source, while top-20 feasible coverage falls sharply at 96.
The cube state target behaves differently because its useful candidates appear
later and across a broader range of proposal scales.

\begin{table}[!ht]
\centering
\scriptsize
\setlength{\tabcolsep}{1.6pt}
\begin{tabular}{llrrrr}
\toprule
Target & Budget/source & Present & Top-20 & Top-1 & Blockers\\
\midrule
Position & 8  & .906 & .906 & .469 & 2.03\\
Position & 24 & 1.000 & .969 & .375 & 4.31\\
Position & 96 & 1.000 & .656 & .062 & 16.91\\
\midrule
Position + yaw & 8  & .906 & .906 & .500 & 1.48\\
Position + yaw & 24 & 1.000 & .969 & .344 & 4.34\\
Position + yaw & 96 & 1.000 & .625 & .031 & 18.31\\
\midrule
Cube & 8  & .625 & .562 & .062 & 7.70\\
Cube & 24 & .969 & .656 & .062 & 15.97\\
Cube & 96 & 1.000 & .562 & .094 & 29.81\\
\bottomrule
\end{tabular}
\caption{Independent overgeneration replication on held-out proposal pools.
Blockers are infeasible candidates ranked below the lowest cost of any feasible
candidate.}
\label{tab:overgen_all}
\end{table}

\begin{table}[H]
\centering
\scriptsize
\setlength{\tabcolsep}{1.6pt}
\begin{tabular}{llrrrr}
\toprule
Split & Target & Peak budget & Peak top-20 & Full top-20 & Full blockers\\
\midrule
Held out A & Position & 24 & .875 & .500 & 23.66\\
Held out A & Position + yaw & 24 & .844 & .594 & 20.38\\
Replication B & Position & 24 & .969 & .656 & 16.91\\
Replication B & Position + yaw & 24 & .969 & .625 & 18.31\\
\bottomrule
\end{tabular}
\caption{Replication of the non-monotone top-20 feasible coverage pattern. Feasible presence is already one at the budget with highest top-20 coverage.}
\label{tab:overgen_replication}
\end{table}

\FloatBarrier
\subsection{Residual Lower Tail Errors under \trm{}}

The independent candidate execution set is rescored with the temporal head and the matched
shuffled label head.  Feasibility labels continue to come only from executing
the fixed candidates.  Table~\ref{tab:trm_overgeneration_full} reports the two
target families used in the main paper.  Temporal \trm{} retains more useful
candidates in the top 20 at the largest budget than latent cost or the shuffled head,
but its blocker count still grows by roughly ten per pool.
The executed top-1 rates from 72 to 288 total proposals fall from .375 to .062
and from .344 to .031 under latent cost.  Under the reachability cost, they fall from .281 to
.188 and from .281 to .156.

\begin{table}[!ht]
\centering
\scriptsize
\setlength{\tabcolsep}{1.4pt}
\begin{tabular}{llrrrrrr}
\toprule
& & \multicolumn{3}{c}{Top-20 exposure} & \multicolumn{3}{c}{Mean blockers}\\
\cmidrule(lr){3-5}\cmidrule(lr){6-8}
Target & Cost & 24 & 72 & 288 & 24 & 72 & 288\\
\midrule
Position & Latent & .906 & .969 & .656 & 2.03 & 4.31 & 16.91\\
Position & Reachability & .875 & .875 & .750 & 2.55 & 6.66 & 16.66\\
Position & Shuffled reach. & .875 & .844 & .625 & 3.97 & 9.16 & 22.03\\
\midrule
Position + yaw & Latent & .906 & .969 & .625 & 1.48 & 4.34 & 18.31\\
Position + yaw & Reachability & .875 & .906 & .719 & 2.79 & 7.59 & 17.75\\
Position + yaw & Shuffled reach. & .875 & .875 & .594 & 4.34 & 6.94 & 23.41\\
\bottomrule
\end{tabular}
\caption{Proposal scaling under latent, reachability, and shuffled label reachability costs. Column headings give total proposals across three sources. Each row contains 32 independently executed pools.}
\label{tab:trm_overgeneration_full}
\end{table}

\section{Complete Closed-Loop Cube Results}

\subsection{Matched Evaluation of Terminal Costs and Output Rules}

Table~\ref{tab:factorial_full} reports all six controllers.  Kernel \asar{} has
the highest event completion success under both terminal costs at every tested
budget.  Reachability cost improves minimum cost selection at 72 and 144
proposals, while the two rules based on minimum cost are equal at 288.

\begin{table}[!ht]
\centering
\small
\setlength{\tabcolsep}{5pt}
\resizebox{\columnwidth}{!}{%
\begin{tabular}{rlrr}
\toprule
Total proposals & Output rule & Latent cost & Reachability cost\\
\midrule
72 & Minimum cost & .067 & .147\\
72 & Least isolated & .187 & .187\\
72 & Kernel \asar{} & \textbf{.347} & \textbf{.333}\\
\midrule
144 & Minimum cost & .053 & .133\\
144 & Least isolated & .213 & .227\\
144 & Kernel \asar{} & \textbf{.293} & \textbf{.333}\\
\midrule
288 & Minimum cost & .133 & .133\\
288 & Least isolated & .147 & .213\\
288 & Kernel \asar{} & \textbf{.320} & \textbf{.307}\\
\bottomrule
\end{tabular}}
\caption{Event completion success for the matched terminal cost and output rule comparison. Each budget uses the same 75 queries and 44 evaluation seeds.}
\label{tab:factorial_full}
\end{table}

\begin{table}[!ht]
\centering
\scriptsize
\setlength{\tabcolsep}{3pt}
\begin{tabular}{rL{0.62\columnwidth}r}
\toprule
Total & Contrast & $\Delta$ [95\% CI]\\
\midrule
72 & Reachability minimum minus latent minimum & +.080 [.000,.156]\\
72 & Kernel ASAR with latent cost minus matching minimum & +.280 [.160,.400]\\
72 & Kernel ASAR with reachability cost minus matching minimum & +.187 [.096,.280]\\
72 & Kernel ASAR with latent cost minus matching least isolated & +.160 [.054,.264]\\
72 & Kernel ASAR with reachability cost minus matching least isolated & +.147 [.039,.250]\\
\midrule
144 & Reachability minimum minus latent minimum & +.080 [.000,.156]\\
144 & Kernel ASAR with latent cost minus matching minimum & +.240 [.143,.342]\\
144 & Kernel ASAR with reachability cost minus matching minimum & +.200 [.107,.301]\\
144 & Kernel ASAR with latent cost minus matching least isolated & +.080 [-.043,.200]\\
144 & Kernel ASAR with reachability cost minus matching least isolated & +.107 [.013,.211]\\
\midrule
288 & Reachability minimum minus latent minimum & +.000 [-.099,.100]\\
288 & Kernel ASAR with latent cost minus matching minimum & +.187 [.068,.312]\\
288 & Kernel ASAR with reachability cost minus matching minimum & +.173 [.068,.271]\\
288 & Kernel ASAR with latent cost minus matching least isolated & +.173 [.066,.290]\\
288 & Kernel ASAR with reachability cost minus matching least isolated & +.093 [-.030,.203]\\
\bottomrule
\end{tabular}
\caption{Paired event completion success contrasts with 95\% bootstrap intervals clustered by seed.}
\label{tab:factorial_contrasts}
\end{table}

Relative to reachability selection by minimum cost, Kernel records 15/1, 16/1,
and 16/3 paired rescues/losses as the proposal budget grows.  Relative to the
reachability least isolated rule, the counts are 15/4, 10/2, and 14/7.  The
terminal cost by Kernel interaction
is $-.093$ $[-.189,.000]$, $-.040$ $[-.146,.068]$, and $-.013$
$[-.179,.139]$.  We therefore report the paired Kernel gain under each cost
directly rather than infer a superadditive interaction.

The matched shuffled label head provides a second control at 288 proposals.
Its event completion success under minimum cost selection is .187 and Kernel reaches .240, a paired gain
of $+.053$ $[-.053,+.167]$ with 11 rescues and 7 losses.  Temporal \trm{} at
the same budget gives a Kernel gain of $+.173$ $[+.068,+.271]$.  The shuffled
head does not reproduce the positive interval obtained with temporal
supervision.

\begin{table}[!ht]
\centering
\small
\resizebox{\columnwidth}{!}{%
\begin{tabular}{lrrrr}
\toprule
Method & Success & Wins/Losses & Cube pos. & Replans\\
\midrule
Latent minimum & .723 & -- & .1172 & 2.95\\
Least isolated & .744 & 17/6 & .1161 & 2.82\\
Kernel \asar{} & .762 & 26/6 & .1061 & 2.70\\
Portfolio \asar{} & \textbf{.766} & \textbf{27/5} & .1064 & \textbf{2.69}\\
\bottomrule
\end{tabular}}
\caption{Cube evaluation over 64 seeds and 512 paired queries. Wins and losses compare each method with selection by minimum latent cost.}
\label{tab:cube_main_full}
\end{table}

The success intervals clustered by seed relative to selection by minimum latent cost are [.0039,.0391] for least isolated selection, [.0156,.0645] for Kernel \asar{}, and [.0215,.0664] for Portfolio \asar{}. Kernel ASAR and Portfolio ASAR also have positive cube position and target position intervals. Exact values are retained in the released bootstrap summary.

\FloatBarrier
Kernel ASAR and Portfolio ASAR are close but not identical. Kernel ASAR succeeds on 26 queries that fail under selection by minimum latent cost and fails on 6 queries that this baseline solves. Portfolio ASAR succeeds on 27 queries that fail under selection by minimum latent cost and fails on 5 queries that this baseline solves. In direct comparison with Kernel ASAR, Portfolio ASAR succeeds alone on 10 queries and fails alone on 8. Their success union is .781. Held-out scalar routers did not reliably realize this union, so the paper reports fixed reconstruction variants rather than a learned test set switch.

\subsection{Controls for Reconstruction from the Final Proposal Pool}

This evaluation holds the model, proposal pools, execution budget, and seven output rules fixed over 64 new seeds and 512 paired queries. It tests whether adjacent-set reconstruction improves beyond generic averaging of low cost candidates.

\begin{table}[H]
\centering
\scriptsize
\setlength{\tabcolsep}{3pt}
\resizebox{\columnwidth}{!}{%
\begin{tabular}{lrrrr}
\toprule
Method & Success & $\Delta$ minimum & W/L minimum & Cube position improvement\\
\midrule
Latent minimum & .7109 & -- & -- & --\\
Latent elite mean & .7207 & +.0098 & 15/10 & .0054\\
Elite mean with minimum anchor & .7168 & +.0059 & 11/8 & .0067\\
Adjacent prefix mean & .7324 & +.0215 & 21/10 & .0078\\
Anchored adjacent prefix mean & .7344 & +.0234 & 21/9 & .0070\\
Kernel \asar{} & \textbf{.7422} & \textbf{+.0312} & \textbf{24/8} & \textbf{.0116}\\
Portfolio \asar{} & .7383 & +.0273 & 24/10 & .0098\\
\bottomrule
\end{tabular}}
\caption{Final proposal pool reconstruction controls on 512 paired queries. Continuous improvement is the error under selection by minimum latent cost minus the method error, so positive is better.}
\label{tab:final_pool_reconstruction}
\end{table}

\begin{table}[H]
\centering
\scriptsize
\setlength{\tabcolsep}{2pt}
\resizebox{\columnwidth}{!}{%
\begin{tabular}{lccc}
\toprule
Comparison & Success & Cube pos. & Target pos.\\
\midrule
vs.\ latent minimum & .0312 [.0098,.0547] & .0116 [.0069,.0167] & .0106 [.0061,.0151]\\
vs.\ latent elite mean & .0215 [.0000,.0430] & .0063 [.0019,.0106] & .0032 [.0003,.0062]\\
vs.\ anchored elite mean & .0254 [.0059,.0449] & .0049 [.0010,.0089] & .0044 [.0013,.0074]\\
\bottomrule
\end{tabular}}
\caption{Kernel ASAR paired differences with 95\% intervals clustered by seed. The success interval versus elite mean under latent cost touches zero. Both continuous error intervals are positive.}
\label{tab:final_pool_pairwise}
\end{table}

Latent cost elite averaging gains roughly one point, while the Kernel ASAR gain over selection by minimum latent cost is 3.1 points. Kernel ASAR also leads elite averaging under latent cost by 2.15 points, with an interval whose lower endpoint is zero, and improves both continuous errors with positive intervals. We use Kernel \asar{} as the main reconstruction variant and Portfolio \asar{} as a robustness variant.

\FloatBarrier
\subsection{Carry and Release Phase Analysis}

The second 512-query evaluation also supports an analysis by physical task phase. The subset rules were recorded before the outcomes were inspected and use only the expert segment associated with each start and goal. The main outcome is event completion success. It requires environment success and a final gripper contact value within .1 of the goal contact value.

\begin{table}[H]
\centering
\scriptsize
\setlength{\tabcolsep}{3pt}
\begin{tabular}{L{0.27\columnwidth}L{0.65\columnwidth}}
\toprule
Query subset & Expert segment rule\\
\midrule
Release transition & Start contact above .5 and goal contact below .1\\
Carry and Release & Release transition with cube height decreasing by more
than 3 cm\\
Full pick and place & Start and goal height below 4 cm, maximum height
above 7 cm, contact above .5, and XY displacement above 3 cm\\
\bottomrule
\end{tabular}
\caption{Phase definitions specified independently of outcomes. Carry and Release begins from an already grasped cube and ends at a released goal.}
\label{tab:carry_release_rules}
\end{table}

\begin{table*}[tbp]
\centering
\scriptsize
\setlength{\tabcolsep}{4pt}
\begin{tabular}{llrrrr}
\toprule
Query subset & Method & (N) & Event success & Environment success & Target position error\\
\midrule
Release transition & Latent minimum & 91 & .154 & .198 & .186\\
& Latent elite mean & 91 & .242 & .253 & .138\\
& Kernel \asar{} & 91 & \textbf{.308} & \textbf{.352} & \textbf{.128}\\
& Portfolio \asar{} & 91 & .297 & .341 & .134\\
\midrule
Carry and Release & Latent minimum & 75 & .133 & .147 & .191\\
& Latent elite mean & 75 & .213 & .213 & .137\\
& Kernel \asar{} & 75 & \textbf{.320} & \textbf{.347} & \textbf{.125}\\
& Portfolio \asar{} & 75 & \textbf{.320} & .333 & .134\\
\midrule
Full pick and place & Latent minimum & 42 & \textbf{.119} & \textbf{.119} & .184\\
& Latent elite mean & 42 & .071 & .071 & .187\\
& Kernel \asar{} & 42 & .048 & .071 & \textbf{.177}\\
& Portfolio \asar{} & 42 & .024 & .071 & .178\\
\bottomrule
\end{tabular}
\caption{Outcomes by physical task phase on the second 512-query evaluation. Target position error is final object to goal distance and lower is better.}
\label{tab:carry_release_full}
\end{table*}

Kernel ASAR improves Carry and Release event success by .187 over selection by minimum latent cost with an interval of [+.069,+.310] obtained by resampling seeds and 19 rescues against 5 losses. Its environment success gain over elite mean under latent cost is .133 with interval [+.025,+.247]. The broader release transition shows the same direction. Full pick and place queries still require the controller to compose grasp, lift, transport, lower, and release within one start and goal interval. Kernel ASAR reduces target position error there but does not improve event completion success.

\begin{figure*}[tbp]
\centering
\includegraphics[width=0.98\textwidth]{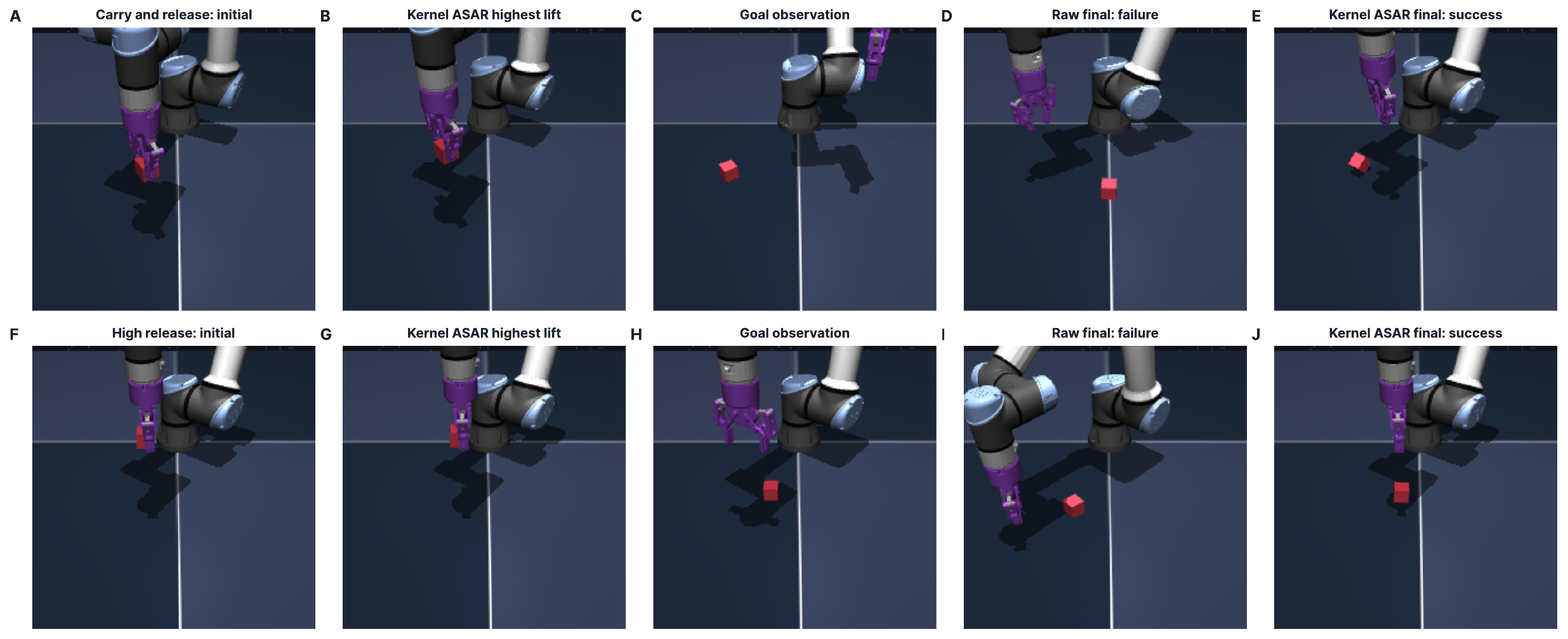}
\caption{Carry and Release executions from the matched paired evaluation. (A-E) A grasped cube is transported to a lower goal. Kernel ASAR lifts the cube, places it near the target, and releases it, while selection by minimum latent cost releases far from the goal. (F-J) A second high release case shows the same physical transition. Every panel is a simulator observation from the executed trajectory.}
\label{fig:carry_release_qualitative}
\end{figure*}

The deterministic cases in Figure~\ref{fig:carry_release_qualitative} provide
the physical interpretation of the final proposal pool aggregate.  They begin after grasp formation,
so the controller must preserve contact during transport and then match a
released goal.  The long horizon boundary is reported separately through the
full pick and place row in Table~\ref{tab:carry_release_full}.

\subsection{Carry and Release Results across Proposal Budgets}

We rerun the same 75 initial states with 24, 48, and 96 proposals per source. All evaluations share the 201,000-row dataset, model checkpoints, metadata for the evaluation queries, controller definitions, and 40-step budget. The row audit matches seed, query identifier, dataset row, episode, start step, and goal step. Pools are regenerated per budget, so the sweep measures closed-loop performance rather than survival within one nested pool.

\begin{table}[tbp]
\centering
\scriptsize
\setlength{\tabcolsep}{3pt}
\begin{tabular}{lrrrr}
\toprule
Budget/source & Latent minimum & Latent elite & Kernel \asar{} & Portfolio \asar{}\\
\midrule
24 & .067 & .147 & \textbf{.360} & .347\\
48 & .053 & .253 & .293 & \textbf{.347}\\
96 & .133 & .213 & \textbf{.347} & .307\\
\bottomrule
\end{tabular}
\caption{Event completion success in matched Carry and Release evaluations across proposal budgets. Total proposal counts are 72, 144, and 288.}
\label{tab:carry_release_budget}
\end{table}

\begin{figure*}[tbp]
\centering
\includegraphics[width=0.88\textwidth]{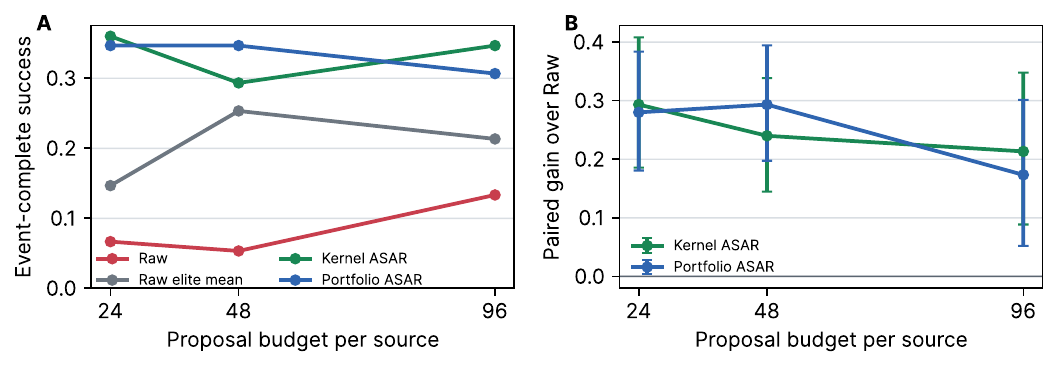}
\caption{Carry and Release performance across proposal budgets. (A) Fixed controller outcomes. (B) Paired gains over selection by minimum latent cost with 95\% intervals clustered by seed. Kernel \asar{} improves that baseline by .293, .240, and .213 as the total proposal count increases from 72 to 288.}
\label{fig:carry_release_budget}
\end{figure*}

Kernel ASAR improves selection by minimum latent cost at every budget with intervals [+.186,+.408], [+.145,+.338], and [+.089,+.348]. The selection by minimum latent cost improves from budget 48 to 96 by .080 with interval [+.013,+.153], so this Carry and Release set does not reproduce the degradation in the candidate execution audit. The position and yaw candidate sets provide the direct proposal overgeneration evidence.

\section{Cube Results Stratified by Task Conditions}

We separate static and moving targets to locate the aggregate Cube gain. The strata use fixed start and goal geometry from the evaluation query, without method outcomes in the bin definitions. Confidence intervals resample the 64 seeds as clusters, while ordinary confidence intervals paired at the query level are retained in the released CSV artifacts. All 512 query keys are paired across the four selectors, and every goal remains inside the same 201-step dataset episode at offset 40.

\begin{table}[tbp]
\centering
\scriptsize
\setlength{\tabcolsep}{3pt}
\resizebox{\columnwidth}{!}{%
\begin{tabular}{lrrrrr}
\toprule
Query stratum & \(N\) & Latent minimum & Portfolio & \(\Delta\) & CI clustered by seed\\
\midrule
All queries & 512 & .723 & .766 & +.043 & [+.021,+.066]\\
Static object target & 342 & 1.000 & 1.000 & .000 & [.000,.000]\\
Any object motion & 170 & .165 & .294 & +.129 & [+.063,+.195]\\
Smaller moving half & 85 & .294 & .588 & +.294 & [+.176,+.411]\\
Larger moving half & 85 & .035 & .000 & -.035 & [-.076,.000]\\
Stable target contact & 400 & .897 & .905 & +.007 & [-.005,+.020]\\
Changed target contact & 112 & .098 & .268 & +.170 & [+.087,+.248]\\
\bottomrule
\end{tabular}}
\caption{Post hoc Cube stratification using query geometry defined independently of outcomes.  Static means object displacement at most \(10^{-8}\).  The 170 moving
queries are split at their median XY displacement, 0.1200.}
\label{tab:cube_query_strata}
\end{table}

The stratified results show that the aggregate gain comes from the 170 object motion queries rather than the 342 static queries, which both methods solve. All 27 successes unique to Portfolio ASAR and all 5 successes unique to selection by minimum latent cost occur in the object motion group. Goals with changed contact show the clearest benefit. The larger translation half is a difficult regime: selection by minimum latent cost succeeds on 3/85 and Portfolio ASAR on 0/85, although Portfolio ASAR still improves mean cube position, target position, and yaw errors by .0159, .0036, and .1403. Thus adjacent-set reconstruction often makes progress on long moves but does not provide the missing final placement behavior needed to cross the success threshold.

\paragraph{Targeted Reconstruction Evaluation.}
The broader contact motion evaluation did not separate action reconstruction from
least isolated existing sequence selection.  We therefore recorded a more
selective rule before a second evaluation: nonzero yaw change, moderate nonzero
object translation, and replan-0
\texttt{kernel\_max\_weight} at most 0.1470682648.  This selects
cases where the reconstruction weight is spread across several candidates
rather than concentrated on one neighbor.  On 96 new seeds the rule selects 73
queries across 54 seeds.

\begin{table}[tbp]
\centering
\scriptsize
\setlength{\tabcolsep}{3pt}
\begin{tabular}{lrrrr}
\toprule
Method & Success & Cube pos. & Target pos. & Yaw\\
\midrule
Latent minimum & .425 & .167 & .090 & .844\\
Least isolated & .493 & .154 & .071 & .843\\
Latent elite mean & .466 & .152 & .062 & .868\\
Elite mean with minimum anchor & .507 & .145 & .059 & .816\\
Adjacent prefix mean & .603 & .127 & .050 & .614\\
Anchored adjacent prefix mean & .603 & .136 & .053 & .737\\
Kernel \asar{} & .630 & .135 & .047 & .762\\
Portfolio \asar{} & \textbf{.658} & \textbf{.128} & .048 & .665\\
\bottomrule
\end{tabular}
\caption{Targeted reconstruction evaluation. The subset rule uses only start and goal geometry and a planner trace feature from the first replanning step, recorded before outcome inspection. Kernel \asar{} improves over least isolated existing sequence selection by .137 in success, with an interval of [+.019,+.238] obtained by resampling seeds, and reduces all three continuous errors.}
\label{tab:targeted_reconstruction_results}
\end{table}

\subsection{Cube Contact Transition Visualization}

The cube contact transition example provides a visualization from the targeted reconstruction setting. The object target requires nonzero translation and yaw change. The initial end-effector contact configuration can also make a single early commitment unreliable. The displayed proposal set uses the same 96 sequences from each of the three fixed proposal scales. Gray curves are cumulative commanded XY action traces. The latent cost minimum is red, the adjacent set is green, and the full sequence reconstructed by Kernel \asar{} is blue. Camera images and top views come from simulator execution rather than the latent rollout.

The preselected example set contains six fixed examples. Three are failures under selection by minimum latent cost repaired by Kernel \asar{} in the targeted reconstruction evaluation set. One is an agreement success, one is a long motion case where all reconstruction variants fail, and one comes from the independent overgeneration replication. Selection within each category follows deterministic geometry and error improvement rules. The accompanying video shows every case with the candidate trajectories, execution under selection by minimum latent cost, adjacent prefix set, and Kernel \asar{} execution under the same initial state.

\begin{figure*}[tbp]
\centering
\includegraphics[width=0.98\textwidth]{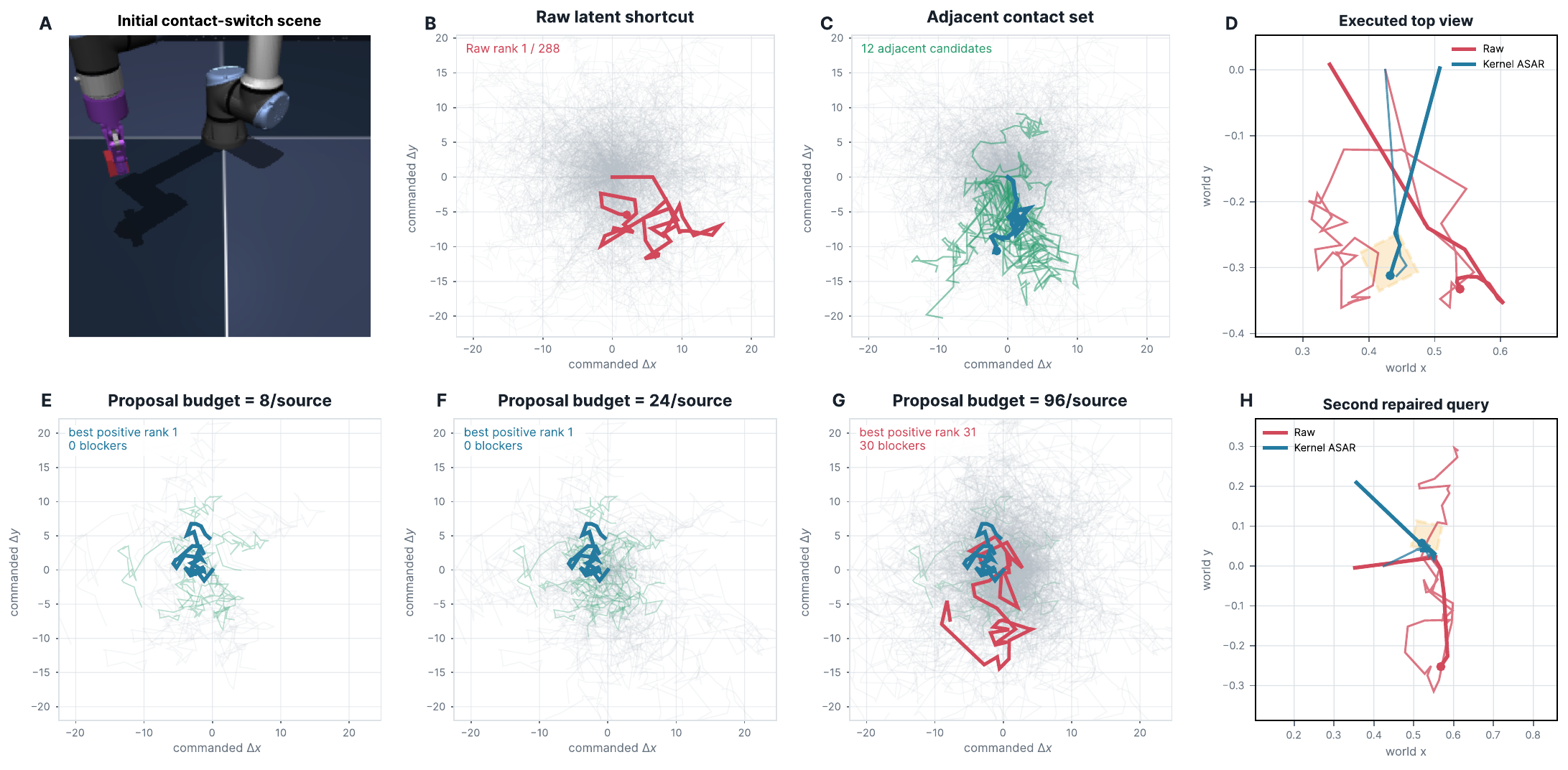}
\caption{Cube contact transition visualization. (A-D) A failure under selection by minimum latent cost repaired by Kernel \asar{}. The minimum cost sequence has an isolated prefix, and the executed top view shows the resulting paths. (E-G) The predetermined proposal budget case from the independent replication. The best feasible position and yaw sequence has latent cost rank 1 at 8 and 24 candidates per source, then falls to rank 31 behind 30 blockers at 96 candidates per source. (H) A second failure under selection by minimum latent cost repaired by Kernel \asar{}.}
\label{fig:contact_switch_qualitative}
\end{figure*}

\paragraph{Reconstruction Gains and Mixture of Controllers Verification.}
Kernel ASAR and Portfolio ASAR succeed on partly different queries. However, a router based only on start and goal geometry and reconstruction traces available before execution does not reliably choose between them. On all 768 new queries, leave-one-seed-out routing collapses to fixed Kernel ASAR (.7448 versus .7396 for Portfolio ASAR); on the 73-query targeted set, it trails Portfolio ASAR (.6301 versus .6575).

\begin{table}[tbp]
\centering
\scriptsize
\setlength{\tabcolsep}{3pt}
\begin{tabular}{lrrrr}
\toprule
Protocol & Success & \(\Delta\) & CI & Branches\\
\midrule
Portfolio first & .7279 & -- & -- & 1.00\\
\moc{} & .7565 & +.0286 & [+.0169,+.0417] & 1.77\\
\bottomrule
\end{tabular}
\caption{\moc{} Validation of the Mixture of Controllers over 96 new seeds and 768 queries. The comparison is against the first branch, Portfolio ASAR in the same sequential evaluator.
\moc{} rescues 22 Portfolio ASAR failures, loses none, and has a 95th percentile of 4.0 executed branches.}
\label{tab:sequential_slate}
\end{table}

Successful branches are distributed as 559 Portfolio ASAR, 12 Kernel ASAR, 8 anchored adjacent prefix mean, and 2 adjacent prefix mean. The remaining 187 queries have no successful branch among the four controllers. The 22 rescues span 19 distinct seeds. A fixed sequence of four controllers turns this complementarity into a 2.86-point gain when verification with reset is available.

\section{Replication with an Independently Trained World Model}
\label{sec:cross_model}

This experiment changes the world model weights and keeps the Cube
environment, query sampler, action normalization, proposal generator,
execution budget, and \asar{} parameters unchanged.  The model is trained with seed
3072 for 100,000 optimization steps on balanced samples from single, double,
triple, quadruple, and octuple Cube play data.  We use the final available
epoch, 32, rather than selecting a checkpoint from control outcomes.  Its
SHA-256 is \texttt{7c0cba50f432a279a7ec03ad20e836d6}\allowbreak
\texttt{e5181e5be9ceae476f8a39995814a9c4}.
Evaluation uses the separate \texttt{ogbench/cube\_single\_expert}
dataset~\citep{park2024ogbench}.

The eligibility pilot contains eight fixed seeds and 64 queries at each budget. Selection by minimum latent cost has 47 successes and 17 failures at 288 total proposals, and Kernel ASAR reconstruction is active in every pool. The predefined evaluation then uses 64 disjoint seeds and eight queries per seed. Both budgets contain the same 512 queries with identical start and goal states. Each source contributes 24 or 96 proposals, giving 72 or 288 proposals in total.

\begin{table}[tbp]
\centering
\scriptsize
\setlength{\tabcolsep}{3pt}
\begin{tabular}{lrrr}
\toprule
Total proposals & Latent minimum & Least isolated & Kernel \asar{}\\
\midrule
72  & .703 & .717 & \textbf{.738}\\
288 & .701 & .713 & \textbf{.727}\\
\bottomrule
\end{tabular}

\vspace{4pt}
\begin{tabular}{rlll}
\toprule
Total & Contrast & $\Delta$ [95\% CI] & W/L\\
\midrule
72 & Kernel minus latent minimum & +.035 [+.014,+.057] & 21/3\\
72 & Kernel minus least isolated & +.021 [+.010,+.035] & 11/0\\
288 & Kernel minus latent minimum & +.025 [+.006,+.047] & 20/7\\
288 & Kernel minus least isolated & +.014 [-.004,+.031] & 13/6\\
\bottomrule
\end{tabular}
\caption{Replication with independently trained world model weights over 64 seeds and 512 paired queries per budget. Intervals resample evaluation seeds 10,000 times.}
\label{tab:cross_model}
\end{table}

At 72 proposals, reconstruction has positive intervals against both predefined baselines. Its margin over least isolated selection shows that returning one existing sequence is insufficient on this evaluation. Kernel ASAR also improves target position error over least isolated selection by .0034 [.0017,.0052]. Cube position error is unchanged and yaw favors least isolated selection, locating the gain in success and target placement rather than every state coordinate.

At 288 proposals, Kernel ASAR remains better than selection by minimum latent cost in success, cube position error, target position error, and yaw error. Relative to least isolated selection, its success interval crosses zero, while all three continuous error improvements remain positive. Success under selection by minimum latent cost remains stable across budgets, changing from .703 to .701 with a paired difference between the large and small budgets of -.002 [-.020,.016]. Adjacent-set reconstruction therefore transfers across independently trained weights. The different budget trends across checkpoints suggest that the effect of proposal scaling depends on the model's residual prediction errors.

The audit matches all 64 predefined evaluation seeds to the recorded set. Both budgets have 512 identical pool keys and zero mismatches in dataset row, episode, start, goal, seed, or query identifier. Every pool contains one output from selection by minimum latent cost, one from least isolated selection, and one from Kernel ASAR. All 128 run records contain the same checkpoint path and hash, and all reported metrics are finite.

\section{Joint Latent and Action Neighborhood Results and Controls}

This section tests whether a joint latent and action neighborhood can preserve a coherent proposal cluster before Portfolio ASAR reconstruction. The first evaluation is used to develop the joint neighborhood rule. The second applies the recorded rule together with random filtering controls matched to the retention rate.

\begin{table}[tbp]
\centering
\scriptsize
\resizebox{\columnwidth}{!}{%
\begin{tabular}{lrrrrrr}
\toprule
Method & Success & \(\Delta\) latent min. & W/L & Cube pos. & Target pos. & Yaw\\
\midrule
Latent minimum & .6797 & .0000 & -- & .0000 & .0000 & .0000\\
Latent elite mean & .7188 & .0391 & 12/2 & .0142 & .0134 & .0716\\
Anchored adjacent prefix mean & .7227 & .0430 & 12/1 & .0123 & .0130 & .0648\\
Portfolio ASAR & .7188 & .0391 & 11/1 & .0121 & .0114 & .0723\\
Latent minimum after graph filtering & .6836 & .0039 & 6/5 & -.0045 & .0006 & -.0629\\
Anchored mean after graph filtering & .7109 & .0312 & 10/2 & .0077 & .0103 & .0229\\
Graph-filtered Portfolio ASAR & \textbf{.7305} & \textbf{.0508} & 15/2 & .0091 & .0096 & .0435\\
\bottomrule
\end{tabular}}
\caption{Development evaluation of joint neighborhood filtering with 256 paired queries.}
\label{tab:graph_blocks}
\end{table}

\begin{table}[tbp]
\centering
\scriptsize
\resizebox{\columnwidth}{!}{%
\begin{tabular}{lrrrrrr}
\toprule
Method & Success & \(\Delta\) latent min. & W/L & Cube pos. & Target pos. & Yaw\\
\midrule
Latent minimum & .7344 & .0000 & -- & .0000 & .0000 & .0000\\
Anchored adjacent prefix mean & .7344 & .0000 & 5/5 & .0083 & .0055 & .0528\\
Portfolio ASAR & .7383 & .0039 & 6/5 & .0055 & .0041 & .0433\\
Graph-Filtered Portfolio ASAR & \textbf{.7422} & \textbf{.0078} & 6/4 & .0066 & .0032 & .0743\\
Portfolio ASAR with random filtering & .7305 & -.0039 & 5/6 & .0042 & .0020 & .0510\\
Matched random & .7305 & -.0039 & 5/6 & .0054 & .0024 & .0611\\
\bottomrule
\end{tabular}}
\caption{Confirmatory evaluation of joint neighborhood filtering and matched controls with 256 paired queries.}
\label{tab:graph_blocks_replication}
\end{table}

\begin{table}[tbp]
\centering
\scriptsize
\begin{tabular}{lrrrr}
\toprule
Filter & Latent filtered & Size & Latent core & Selected core\\
\midrule
Graph component & .6001 & 40 & 3.8432 & .5797\\
Random component & .4995 & 40 & 3.8942 & 2.1078\\
Matched random & .5933 & 40 & 3.8842 & 2.0743\\
\bottomrule
\end{tabular}
\caption{Control for joint neighborhood filtering. The random filter matched to the retention rate reproduces the filtering rate but not joint neighborhood centrality or the associated success gain.}
\label{tab:graph_mechanism}
\end{table}

\FloatBarrier
Filtering to the graph core is effective when coupled to adjacent-set reconstruction on
the development evaluation.  On the second evaluation, latent minimum success
is .7344, Portfolio ASAR .7383, Graph-Filtered Portfolio ASAR .7422, and both random
controls .7305.  The graph increment over Portfolio ASAR is \(+.0039\),
so graph projection is a proposal cluster diagnostic rather than the primary
method.

\begin{table}[tbp]
\centering
\scriptsize
\setlength{\tabcolsep}{2.5pt}
\begin{tabular}{lcc}
\toprule
Pairing and feature & Positive gap & Fixed-\(\beta\) coverage\\
\midrule
Failed latent minimum, Prefix-1 & .548 [.387,.710] & .419 [.258,.581]\\
Failed latent minimum, terminal latent & .742 [.581,.871] & .516 [.355,.677]\\
Failed latent minimum, full sequence & .484 [.323,.645] & .323 [.161,.484]\\
Failed latent minimum, rolled prefix & .548 [.387,.710] & .194 [.065,.323]\\
\midrule
Least isolated feasible, Prefix-1 & .774 [.613,.903] & .613 [.452,.774]\\
Least isolated feasible, terminal latent & .677 [.516,.839] & .419 [.258,.581]\\
Least isolated feasible, full sequence & .645 [.484,.806] & .484 [.323,.645]\\
Least isolated feasible, rolled prefix & .581 [.419,.742] & .258 [.129,.419]\\
\bottomrule
\end{tabular}
\caption{Prefix density audit on held out data at the candidate level. Each row has 31 pairs, and brackets report 95\% intervals.}
\label{tab:gap_audit}
\end{table}

\section{Prefix Density Audit at the Candidate Level}
\label{sec:gap_audit}

The audit instantiates Proposition 4 using executed candidate pools.  The
ranking blocker is the failed candidate with minimum latent cost inside the 80
lowest cost candidates.  The primary feasible candidate is the successful
candidate with smallest latent cost in the same pool.  The secondary feasible
candidate is the successful candidate with the smallest prefix isolation score.
The score uses prefix length 1,
\(k=3\), robustly standardized kNN distance.  The calibration coefficient is
selected on discovery seeds as the 0.75 quantile of the required correction
and then fixed for held-out seeds and controls.

For the latent minimum pairing, the positive prefix gap rate is .548 and the
random assignment test within the pool gives \(p=.3605\).  Fixed-\(\beta\) coverage
is .419 with \(p=.2585\).  For the pairing with the least isolated feasible candidate,
these values rise to .774 (\(p=.0032\)) and .613 (\(p=.0002\)).  Successful
low isolation neighborhoods are therefore common, while the closest successful
candidate in latent cost order need not be less isolated than the blocker.

This distinction matches the method design.  \asar{} reconstructs a full action sequence anchored to the minimum cost sequence and formed from an adjacent set. It does not rely on scalar pairwise correction alone, and Proposition 4 explains only the reranking part of the mechanism.

\section{Reacher Transfer Evaluation}

The transfer test across tasks uses the official Reacher checkpoint and
\texttt{dmc/reacher\_random}.  The Cube Portfolio parameters are used unchanged:
80 lowest cost candidates, prefix length 1, \(k=3\), elite size 7, ratio
thresholds 1.0/1.75, and anchor weights on the sequence with minimum cost of 0.25/0.15/0.10.  No Reacher
threshold or prefix density
parameter is tuned.  Each of five seeds evaluates 50 unique paired episodes
with goal offset 25, budget 50, and \cem{} configured with 300 samples, 30
iterations, and 30 elites.
The supplementary \texttt{prefix\_roll1} control cyclically permutes prefix
feature rows before computing KNN isolation.  It preserves the candidate pool
and computation while breaking the alignment between each candidate and its
prefix features.

\begin{table}[!ht]
\centering
\scriptsize
\setlength{\tabcolsep}{3pt}
\begin{tabular}{lrrr}
\toprule
Method & Success (rate) & \(\Delta\) & W/L\\
\midrule
CEM mean & 200/250 (.800) & .000 & --\\
Latent minimum & 199/250 (.796) & -.004 & 28/29\\
Control with permuted prefixes & 198/250 (.792) & -.008 & 26/28\\
Portfolio ASAR & 199/250 (.796) & -.004 & 21/22\\
\bottomrule
\end{tabular}

\vspace{3pt}
\begin{tabular}{lrr}
\toprule
Method & Episode CI & CI clustered by seed\\
\midrule
CEM mean & [.000,.000] & [.000,.000]\\
Latent minimum & [-.064,.056] & [-.036,.024]\\
Control with permuted prefixes & [-.064,.048] & [-.064,.048]\\
Portfolio ASAR & [-.056,.048] & [-.064,.044]\\
\bottomrule
\end{tabular}
\caption{Reacher transfer with unchanged Cube parameters.  The exact paired
test gives \(p=1.0\) for Portfolio ASAR against CEM mean.}
\label{tab:reacher_main}
\end{table}

\begin{table}[!ht]
\centering
\small
\resizebox{\columnwidth}{!}{%
\begin{tabular}{rrrrr}
\toprule
Seed & CEM mean & Latent min. & Permuted prefix & Portfolio\\
\midrule
42 & .84 & .88 & .76 & .88\\
43 & .76 & .70 & .82 & .78\\
44 & .78 & .78 & .74 & .66\\
45 & .82 & .80 & .76 & .80\\
46 & .80 & .82 & .88 & .86\\
\bottomrule
\end{tabular}}
\caption{Reacher success rates for each seed.}
\label{tab:reacher_seed}
\end{table}

\FloatBarrier
Portfolio ASAR reaches .796 on Reacher compared with .800 for standard CEM mean, with an interval of [-.064,.044] obtained by resampling seeds. The Reacher candidate pools do not yield the Cube improvement under the same prefix geometry. We next test the same reconstruction rule on Push-T, a task with frequent contact and a nonzero fixed \lewm{}
baseline.

\section{Push-T Transfer Evaluation}

Push-T uses the official fixed LeWM checkpoint and
\texttt{pusht\_expert\_train}.  The final CEM pool contains 64 candidates after
10 iterations with eight elites.  The Cube portfolio is copied without tuning:
prefix length 1, \(k=3\), elite size 7, ratio thresholds 1.0/1.75, and
anchor weights on the sequence with minimum cost of 0.25/0.15/0.10.  Three seeds each evaluate 50 paired episodes at goal
offset 50 and budget 50.  The protocol and decision rule were recorded before
the outcomes, and the reported evaluation uses those settings unchanged.

\begin{table}[!ht]
\centering
\scriptsize
\setlength{\tabcolsep}{3pt}
\begin{tabular}{lrrrrr}
\toprule
Cost & Mean & Portfolio ASAR & W/L & Episode CI & Distance \(\Delta\)\\
\midrule
Latent & .273 & .280 & 6/5 & [-.040,.053] & +1.95\\
True state hybrid & .413 & .393 & 6/9 & [-.073,.027] & -5.84\\
\bottomrule
\end{tabular}
\caption{Push-T transfer with unchanged Cube parameters over 150 paired
episodes per cost.  Positive distance \(\Delta\) means Portfolio ASAR reduces final task
distance.}
\label{tab:pusht_main}
\end{table}

Under latent cost, CEM mean reaches .273 and Portfolio ASAR reaches .280 with interval
[-.040,.053].  The three paired seed rates are .30/.30, .22/.32, and .30/.22
for CEM mean/Portfolio ASAR.
Under true state hybrid cost, Portfolio ASAR loses two points in every seed
(.44/.42, .34/.32, and .46/.44) and worsens final distance by 5.84; the
distance interval obtained by resampling episodes is \([-9.72,-2.41]\).
The latent exact sign test \(p\)-value is 1.0 and its success CI clustered by seed is
\([-.080,.100]\); the corresponding true state hybrid values are .607 and
\([-.020,-.020]\).
Portfolio ASAR remains active, using 52 compact, 88 intermediate, and 160 dispersed branches
under true state hybrid cost.  Contact richness alone does not reproduce the Cube
prefix geometry.  A transferred prefix metric must also match the task's
proposal distribution.

\FloatBarrier
\section{Alternative Diagnostics and Controls}

The comparisons in Table~\ref{tab:negative} trace the route from latent
structure to the final planner intervention.  They show which signals are
descriptive, which survive closed-loop evaluation, and which component is
needed to convert prefix density into task success.
The 512-query final proposal pool evaluation gives .711 for latent selection by minimum cost, .721 for latent elite mean, .717 for elite mean with minimum anchor,
.732 for adjacent prefix mean, .734 for anchored adjacent prefix mean, .742 for
Kernel ASAR, and .738 for Portfolio ASAR.  Latent elite averaging has a smaller gain than
Kernel ASAR, although the Kernel ASAR success interval against that baseline touches
zero.

\begin{table}[!ht]
\centering
\scriptsize
\setlength{\tabcolsep}{3pt}
\begin{tabular}{L{0.29\columnwidth}L{0.62\columnwidth}}
\toprule
Alternative & Outcome and interpretation\\
\midrule
Static task rowspace & Variables are decodable, but top-1 success changes only
from .323 to .328.\\
Terminal latent density & Diagnostic structure does not match action prefix
reconstruction in closed-loop control.\\
Index and prefix penalties & Blocker exposure is reproduced without reliable
receding-horizon gain.\\
Latent tangent metrics & Some continuous errors improve, while success trails
Kernel ASAR and Portfolio ASAR.\\
Learned candidate rankers & Gains on held out data vanish under shuffled and
composition controls.\\
Kernel/Portfolio ASAR router & The variants are complementary, but learned switches
do not reliably beat the best fixed method.\\
Graph filtering alone & Selection from the lowest latent costs is nearly flat.  Gains require graph
projection followed by reconstruction.\\
Controls for the minimum cost anchor & Alternative anchors retain a partial gain.  All
effective variants retain adjacent-set reconstruction.\\
\bottomrule
\end{tabular}
\caption{Diagnostic alternatives and their planning outcomes.}
\label{tab:negative}
\end{table}

\FloatBarrier

\section{Implementation and Reproducibility}

The accompanying Code and Data Supplement contains the Cube reconstruction rules, transfer evaluators, analysis scripts, recorded configurations, aggregate results, and representative candidate rows used by this paper. Its README provides a package map for readers, while \texttt{CLAIM\_ARTIFACT\_MAP.md} links the paper claims to the corresponding released records. Exact commands, file paths, seed lists, hashes, and complete software snapshots are kept in that package rather than repeated in the PDF.

Paired comparisons reuse the same query identifiers, initial states, evaluation seeds, and proposal keys required by each protocol. Aggregation checks the expected paired keys and finite numeric fields before computing the reported summaries and confidence intervals. Reacher additionally checks identical dataset rows and solver seeds across methods. Push-T preserves paired episode keys and the branch information recorded at each replan.

The confirmatory Cube evaluations used Ubuntu 22.04.5 on x86-64 hardware with two AMD EPYC 7763 processors and two NVIDIA RTX 6000 Ada GPUs. The recorded environment used CUDA 12.8, Python 3.10.20, and PyTorch 2.10.0 with CUDA 12.8 support. The artifact contains the complete package versions and launch environment for the reported runs.

{\small
\bibliography{references}
}

\end{document}